\documentclass[10pt,twocolumn,letterpaper]{article}
\usepackage{booktabs,multirow,array,tabularx}
\usepackage{xcolor,colortbl}
\usepackage{tcolorbox}
\usepackage{amsmath,amssymb,amsthm}
\newtheorem{theorem}{Theorem}

\newtheorem{proposition}[theorem]{Proposition}
\newtheorem{corollary}[theorem]{Corollary}
\usepackage{graphicx}
\definecolor{coverblue}{HTML}{245E8B}
\definecolor{muted}{HTML}{68737D}
\definecolor{referencebg}{HTML}{EAF3F0}
\definecolor{sectionbg}{HTML}{F0F3F5}

\newtcolorbox{findingsbox}[1]{
    title={#1},
    fonttitle=\bfseries,
    fontupper=\small,
    colback=black!2,
    colbacktitle=black!7,
    coltitle=black,
    colframe=black!30,
    boxrule=0.4pt,
    arc=1pt,
    left=5pt,
    right=5pt,
    top=4pt,
    bottom=4pt,
    before skip=6pt,
    after skip=6pt
}

\usepackage{cvpr}              
\definecolor{cvprblue}{rgb}{0.21,0.49,0.74}
\usepackage[pagebackref,breaklinks,colorlinks,allcolors=cvprblue]{hyperref}

\def\confName{CVPR}
\def\confYear{2027}

\title{On the Limits of Observability for Evaluating Coverage in Machine Unlearning }
\author{%
  \textbf{Ziluowen Luo$^1$}, \textbf{Jun Yin$^2$}, \textbf{Songyang Liu$^3$},  \textbf{Ji Qi$^3$}, \textbf{Chuang Hu$^1$}, \\ 
  \textbf{Kyrie You Wu$^3$}, \textbf{Chaozhuo Li$^3$},  \textbf{Senzhang Wang$^{1}\thanks{Corresponding author}$}\\
    \textsuperscript{1} Central South University, 
    \textsuperscript{2} Hong Kong Polytechnic University \\
    \textsuperscript{3} Beijing University of Posts and Telecommunications \\
    \ttfamily \small
\{lzlwddl, szwang\}@csu.edu.cn, Junmay.yin@connect.polyu.hk, {lichaozhuo1991}@gmail.com
}

\begin{document}
\maketitle

\begin{abstract}
Machine unlearning aims to remove the influence of any requested subset of training data while preserving utility on the remaining data. 
Empirical evaluations measure predictive accuracy and infer training-set membership (MIAs) from model outputs or intermediate activations. 
However, by focusing on selected forget sets, these evaluations can overlook \textbf{Coverage}, the range of requests a method can fulfill. 
To assess this coverage, we first conduct a stress test of \textbf{observability} with 502 retrained models, finding that hidden embeddings reveal known deletions missed in predictions. 
To characterize capacity limits on coverage, we show that distinguishing all $n$ singleton deletions with linear scores can require a shared representation of rank $\Omega(\log n)$. 
We further extend the bound to arbitrary subsets and show when partial deletion can appear complete. 
Building on this analysis, we propose \textbf{CEO}, a framework for evaluating unlearning coverage under limited observability. 
Its compositional color–shape benchmark makes forgetting and retained utility measurable from the full output distribution and provides a leave-one-out retraining oracle for each target. 
From a fixed number of oracle observables, we estimate empirical target coverage, the fraction of targets reliably flagged when deleted and rarely flagged under any other deletions. 
To recover missed deletions, CEO further expands each observable representation into high-dimensional activation space with the frozen encoder of a SAE trained on the original model. 
Extensive experiments of 20 baselines shows that even on this simple synthetic task, only retraining from scratch covers all deletion targets. 
\end{abstract}
\section{Introduction}\label{sec:intro}
\vspace{-0.2cm}


Machine learning models derive their capabilities from training data, whose use raises concerns about privacy, security, and data governance~\cite{luo2026vocabulary,yin2025unleash,luo2026nice,luo2026distilling}. 
These concerns motivate \emph{machine unlearning (MUL)}, which aims to remove the influence of specified training data while preserving model utility~\cite{luo2026otk}. 
Retraining from scratch provides a natural reference for arbitrary subset a deletion request specify, since the resulting model learns from the retain training examples alone~\cite{bourtoule2021machine}. 
In practice, however, an unlearning update procedure may only approximate this outcome, and \emph{unlearning evaluation} aims to assess how closely the resulting model reproduces the behavior of its retrained counterpart.

To avoid full retraining for each deletion request, prior works have explored how to organize training so that unaffected computations can be reused~\cite{ginart2019making,bourtoule2021machine} and how to adapt an already trained model using the forget and retain sets~\cite{kurmanji2023towards,fan2024salun}.
To assess such methods, empirical evaluations typically query the resulting model and examine its prediction outputs or selected intermediate-layer activations, which we collectively term \emph{model observables}. 
For example, researchers use model responses to measure accuracy on the retain set and to test, through membership inference attacks (MIAs), whether forget examples remain distinguishable from nonmembers~\cite{kurmanji2023unbounded}. 
Although these evaluations assess both effectiveness and efficiency, their results characterize a unlearning method on the forget sets chosen for evaluation. 
This leaves open a broader question. 
\textit{Q1: Can an unlearning method delete every requested subset?}

\begin{figure}[t]
    \centering
    \includegraphics[width=1\linewidth]{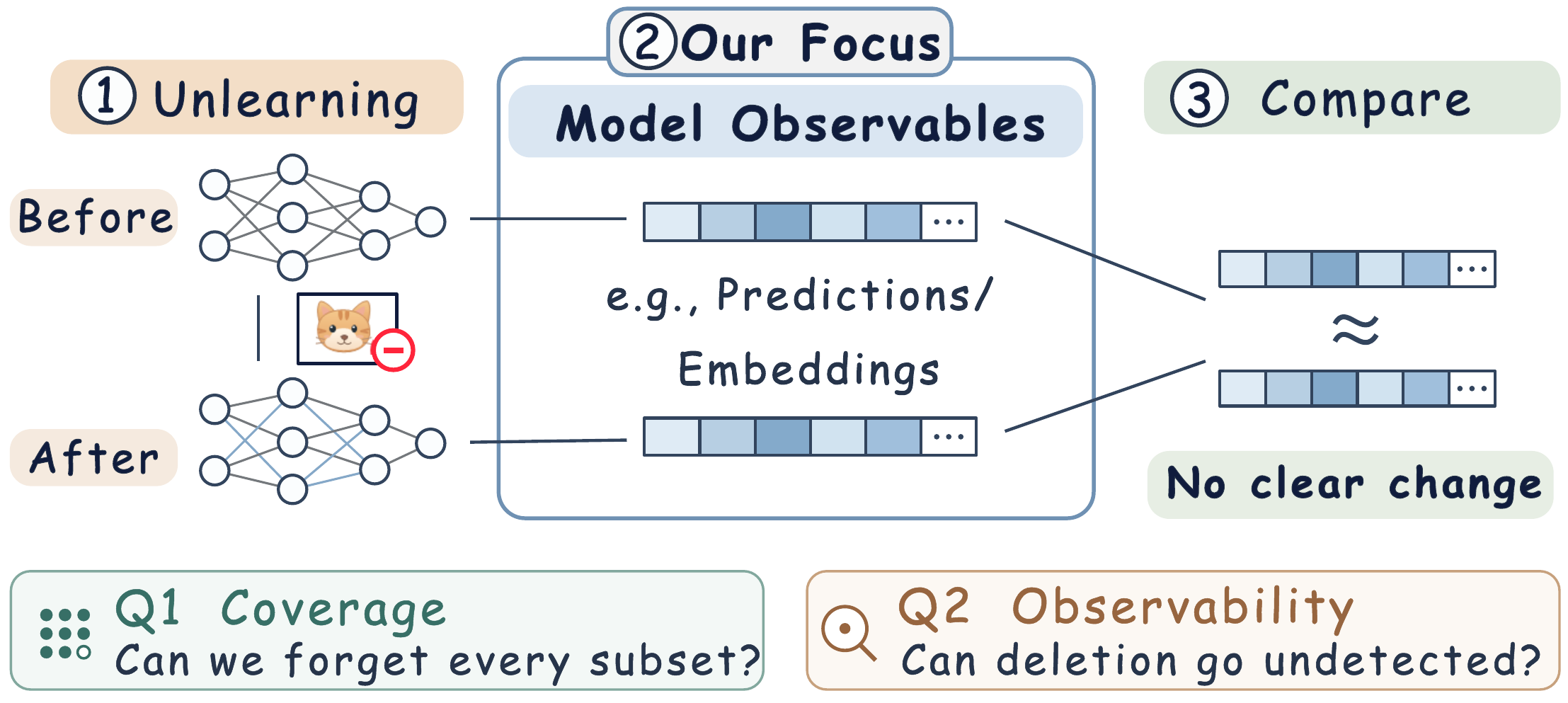}
\caption{\textbf{Assessing coverage relys on observability.}
}
    \label{fig:fig1}
    \vspace{-0.5cm}
\end{figure}

To answer this question, we use \textbf{\emph{Coverage}} to describe the range of requests for which the same unlearning method achieves both forgetting and retained utility. 
Assessing this coverage from model observables requires comparing responses before and after unlearning to gauge the extent of forgetting.  
However, when this comparison shows \textbf{\emph{No Clear Change}} in prediction, the target may still influence the model, or its deletion may have gone undetected. 
To isolate the latter possibility, we use retraining as an oracle, leaving out one target at a time so that its exclusion is known independently of the observed responses (\S~\ref{sec:motivating-observation}). 
For $7.8\%$ of the targets in Figure~\ref{fig:fig2}(b), the evaluator detected forget only when it looked beyond the original classifier-head subspace. 
For the target shown in (c), expanding from 36D to 256D revealed a direction where the oracle's response lay well above the mean response of models that retained the target ($+4.97$). 
The coverage we can establish thus depends on \textbf{\emph{Observability}}, the extent to which the available responses distinguish a target's deletion from outcomes in which it is retained.
\textit{Q2: When can a deletion go undetected, and what additional observables could reveal it?}

To explore how these \emph{blind spots} affect coverage assessment, we study the capacity a shared evaluation representation needs to reveal whether each requested subset was deleted in full. 
Since a \emph{singleton request} can specify any one example, we first require each target's linear score to respond strongly to its deletion and weakly to other deletions. 
Even under oracle retraining, sustaining this contrast can force the rank of the shared representation to grow at least logarithmically with the number of targets. 
We further extend the analysis to \textit{arbitrary subsets}, where linear scores that separate complete deletion from full retention can all fail to sustain that contrast against partial deletions.  
A high score offers little evidence of completion if the evaluator gives the same request a comparable score under partial deletion.
Therefore, we use known deletion outcomes to establish which deletions the evaluator detects and whether additional observables reveal more on the same model pairs.

To \textit{hit two birds with one stone}, we propose \textbf{CEO}, a \underline{C}overage \underline{E}valuation framework for unlearning under limited \underline{O}bservability. 
Specifically, we first design a simple compositional color–shape benchmark in which leave-one-out retraining provides a deletion oracle for testing whether an unlearning method can suppress any requested pair while preserving utility on the remaining pairs. 
We then use these retrained models to measure \textit{empirical target coverage}, the fraction of targets whose deletion we can reliably detect from a fixed number of responses while controlling false positives under other deletions. 
Inspired by the capacity analysis, we fit an overcomplete sparse autoencoder (SAE) on the original model, then expand each observable into high-dimensional activation space with its frozen encoder to detect deletions missed. 
Our contributions are threefold:
\begin{enumerate}[leftmargin=*,noitemsep]
    \item We introduce a novel \textit{coverage} perspective on \textit{unlearning evaluation} that exposes an \textit{observability} gap between the range of the arbitrary requests a method fulfills and the deletions that model observables can reliably reveal.

    \item We establish capacity limits for evaluation on a shared representation, extending our $\Omega(\log n)$ rank bound for covering all $n$ singletons to arbitrary subset requests and showing when partial deletion can appear complete.

    \item We propose \textbf{CEO} to measure \textit{empirical target coverage} by testing which known deletions are detectable. Guided by our capacity bounds, it applies \textit{Dimensional Expansion} to reveal missed deletions from the same responses.

    \item Our experiments show that even on a simple compositional benchmark, few configurations of existing unlearning methods cover every singleton deletion request.
\end{enumerate}
\section{Related Work}\label{sec:work}\noindent\textbf{Machine unlearning.}
Machine unlearning aims to remove the influence of specified training data while preserving model utility~\cite{cao2015towards,luo2026otk,huo2025mmunlearner}. 
Existing approaches reduce deletion costs by organizing training for localized retraining~\cite{bourtoule2021machine}, or modifying trained models through distillation~\cite{kurmanji2023unbounded} and selective parameter updates~\cite{foster2024ssd,fan2024salun}. 
While these efforts pursue effective deletion without full retraining, we revisit the evidence used to judge their success: whether the available model observables can reliably reveal the deletion effects.

\noindent\textbf{Unlearning evaluation.}
Unlearning evaluations commonly combine predictive performance (e.g., Accuracy) on forget and retain data with membership inference tests (e.g., MIA) for residual training-data influence~\cite{kurmanji2023unbounded}.
Retraining-based protocols further assess agreement between the response distributions of unlearned and retrained models~\cite{triantafillou2024competition}.
The reliability of these assessments has also been examined through stronger per-example membership tests~\cite{hayes2025inexact} and analyses of how evaluation procedures can distort conclusions about removal~\cite{feng2025inconclusive}. 
Beyond the choice of evaluation protocol, reliable detection also requires deletion effects to be distinguishable in the available model observables. 
In this work, we focus on how the dimension of these observables constrains evaluation coverage, measured by the fraction of targets whose deletion can be reliably detected.
\section{Problem Setup and Motivation}\label{sec:motivation}


In this section, we first define the unlearning task and adopt retraining on the retain set as the oracle for evaluation (\S~\ref{sec:preliminaries}).
We then conduct an oracle experiment to examine whether an evaluator can fail to distinguish target from non-target deletions even when unlearning is exact (\S~\ref{sec:motivating-observation}). 

\begin{figure*}
    \centering
    \includegraphics[width=1\linewidth]{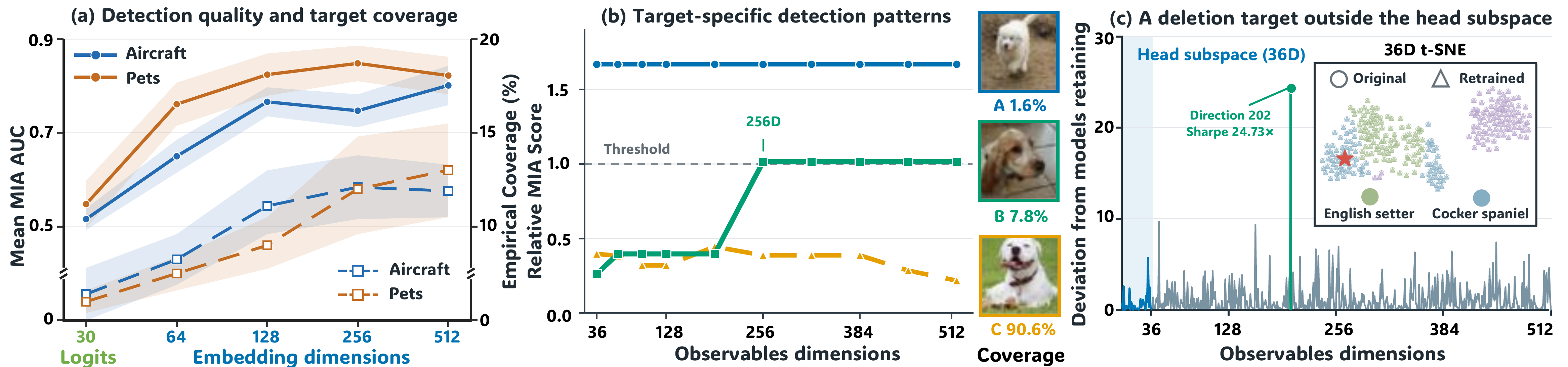}
\caption{\textbf{Evaluating oracle deletions through different observables.}
(a) The curves report mean per-target MIA AUC and coverage rate, respectively. 
(b) Scores are divided by the threshold. \% denote the proportions of different cases A, B and C in dataset.
(c) For example B, a large deviation outside the 36D head subspace drives the 512D space. 
The inset shows little separation between original and retrained.
}
    \label{fig:fig2}
    \vspace{-0.5cm}
\end{figure*}

\subsection{Preliminaries}
\label{sec:preliminaries}


\noindent \textbf{Unlearning task.}
Given a training set $\mathcal D$, a randomized training procedure $\mathcal A$ produces the \emph{original model} $M_{\mathcal D}=\mathcal A(\mathcal D)$. 
Each unlearning request targets a nonempty forget subset $\mathcal F\subset\mathcal D$, possibly a \textit{singleton} ($|\mathcal F|=1$), leaving the retain set $\mathcal R=\mathcal D\setminus\mathcal F$. 
To remove the influence of $\mathcal F$ while preserving utility on the retain set~\cite{bourtoule2021machine}, an unlearning algorithm $\mathcal U$ returns the \emph{unlearned model} $M^{\mathrm u}=\mathcal U(M_{\mathcal D},\mathcal F)$. 

\noindent \textbf{Retraining oracle.}
Exact unlearning requires $M^{\mathrm u}$ to match the distribution of models trained without $\mathcal F$. 
Retraining from scratch on $\mathcal R$ therefore provides the \emph{unlearning oracle}:
\begin{equation}
M_{\mathcal R}=\mathcal A(\mathcal R)
=\mathcal A(\mathcal D\setminus\mathcal F).
\label{eq:training-oracle}
\end{equation}
Despite its cost, we use the oracle to test whether an evaluator can miss a target deletion even under exact unlearning.

\noindent\textbf{Model observables.}
For a trained model $M$ and input $x$, let $z_M(x)\in\mathbb R^p$ denote the selected intermediate-layer activations or prediction outputs, termed a \emph{model observable}.
For the fixed query inputs $\mathcal Q$ and a forget request $\mathcal F$, we stack the paired observables before and after oracle deletion:
\begin{equation}
Z_{\mathcal F}
=
\bigl(z_{M_{\mathcal D}}(x),z_{M_{\mathcal R}}(x)\bigr)_{x\in\mathcal Q}
\in\mathbb R^{2p|\mathcal Q|}.
\label{eq:deletion-observations}
\end{equation}
We use $Z_{\mathcal F}$ as the basis for our subsequent evaluation analysis, with the extraction rule fixed across deletion requests.


\subsection{A Motivating Observation}
\label{sec:motivating-observation}

Existing efforts have made progress in balancing forgetting and retained utility, with the performance assessed through selected model observables on the forget and retain sets. 
However, the evaluations alone do not establish whether the methods remain effective across arbitrary forget subsets. 
To investigate this coverage, we begin with a stress test to distinguish deletion from retention across different requests.

\noindent\textbf{Stress test under an oracle setting.}
Exhaustive evaluation is computationally prohibitive as the number of possible forget subsets grows exponentially with training-set size. 
We therefore focus on singleton requests $\mathcal F=\{x\}$, since support for arbitrary forget subsets must include deleting individual examples. 
Retraining on $\mathcal D\setminus\{x\}$ serves as the \textit{oracle deletion}, excluding unsuccessful unlearning as an explanation for missed detections. 
We then use a membership inference attack (MIA) to distinguish target deletion from retention, summarizing its scores by average AUC and the fraction of targets whose deletions are detected within a prescribed per-target false-positive budget. 
In this experiment, we train 502 models across two datasets to support the following analysis, keeping them fixed while extending evaluation to higher-dimensional intermediate embeddings.

\noindent\textbf{Finding 1: Oracle deletions can remain undetected.}
Figure~\ref{fig:fig2}(a) shows that many oracle deletions remain undetected under the prescribed false-positive budget, even when mean MIA AUC exceeds chance. 
Since each target was excluded by retraining, these misses can not be attributed to unsuccessful unlearning. 
We therefore examine individual targets to determine whether their missed detections persist as additional observables become available.

\noindent\textbf{Finding 2: Additional observables can reveal missed deletions.}
Figure~\ref{fig:fig2}(b) illustrates three patterns: targets detected with the initial observables (A), missed initially but detected in an expanded space (B), and undetected across all tested spaces (C).
For B, detection is recovered at 256 dimensions under the same false-positive budget while all models remain unchanged, demonstrating that the initial miss can be resolved through evaluation alone.
Figure~\ref{fig:fig2}(c) further shows that B's largest standardized deviation from models retaining it lies outside the 36-dimensional head subspace, revealing a signal omitted by the initial observables.
This recovery raises a broader question: \emph{what representational capacity is required to distinguish deletion from retention across arbitrary training targets?}

\section{Capacity Limits of Unlearning Evaluation}\label{sec:diag}

To answer this question, we analyze the evaluation capacity under oracle unlearning, with deletions performed by retraining. 
Since coverage of arbitrary forget subsets requires singleton coverage, we first establish capacity bounds for singleton requests (\S~\ref{sec:singleton-capacity}).
We then extend the analysis to subset requests, where distinguishing complete from partial deletion introduces additional constraints (\S~\ref{sec:subset-capacity}). 
These results further inform design principles for expanding deletion detection and evaluating the target coverage ability .

\subsection{Capacity Limits for Singleton Coverage}
\label{sec:singleton-capacity}

\noindent\textbf{Evaluation framework.}
Common unlearning evaluations, including membership inference attacks (MIAs)~\cite{shokri2017membership,carlini2022membership} and prediction accuracy (ACC), derive their assessments from selected model observables. 
To study singleton coverage within this paradigm, consider $n \geq 2$ targets $u_1,\ldots,u_n\in\mathcal D$ and write $Z_i=Z_{\{u_i\}}$ for the paired observables associated with the oracle deletion of $u_i$. 
We further consider a general evaluator that forms a shared representation $\psi(Z_i)\in\mathbb R^d$ and computes a target-specific score $h_j(Z_i)$ to assess whether the queried target $u_j$ has been deleted. 
We then study the capacity required for this shared representation to support deletion detection for all $n$ targets.

\noindent\textbf{Singleton coverage criterion.}
To assess the capacity, we impose a common selectivity criterion requiring each score to respond strongly to its target's deletion and weakly to other deletions. 
With the oracle models fixed, we analyze linear scores on the shared representation:
\begin{equation}
C_{ij}=h_j(Z_i)=w_j^\top\psi(Z_i),
\qquad
w_j,\psi(Z_i)\in\mathbb R^d.
\label{eq:shared-target-score}
\end{equation}
The map $\psi$ may be nonlinear, and $d$ counts all features supplied to each score.
For target $u_j$, $C_{jj}$ scores its deletion, whereas $C_{ij}$ ($i\neq j$) scores a deletion that retains it.
We call $h_j$ \emph{selective} for the queried target $u_j$ at levels $a>b\geq0$ if
\begin{equation}
C_{jj}\geq a,
\qquad
|C_{ij}|\leq b\quad(i\neq j),
\qquad
\rho=\frac{b}{a}.
\label{eq:target-selectivity}
\end{equation}
Here $\rho$ bounds the relative non-target response, with smaller values enforcing stronger selectivity across requests.


\noindent\textbf{From selectivity to capacity.}
Full singleton coverage requires the same $d$-dimensional shared representation to support selective responses for all $n$ deletion requests.
Let $v_i=\psi(Z_i)$, and stack the representation and weight vectors as columns of $V=[v_1,\ldots,v_n]$ and $W=[w_1,\ldots,w_n]$, both in $\mathbb R^{d\times n}$.
The score matrix then satisfies
\begin{equation}
C=V^\top W\in\mathbb R^{n\times n}, 
\operatorname{rank}(C)\leq r:=\operatorname{rank}(V)\leq d.
\label{eq:shared-response-rank}
\end{equation}
When $\rho=0$, each selective score is positive for its target's deletion and zero otherwise. 
Thus, $C$ is diagonal with rank $n$, implying $d\geq n$ for full singleton coverage. 
The theorem extends the bound to $\rho>0$, allowing non-target responses.

\begin{theorem}[\textbf{Capacity limits for singleton coverage}]
\label{thm:singleton-capacity}
For $n\geq2$, if every target satisfies Equation~\eqref{eq:target-selectivity}, then
\begin{equation}
d\geq r\geq
\frac{n}{1+(n-1)\rho^2}.
\label{eq:singleton-basic-bound}
\end{equation}
Moreover, when $n^{-1/2}\leq\rho<1/2$, a universal constant $c>0$ gives
\begin{equation}
d\geq r\geq
c\,\frac{\log n}{\rho^2\log(1/\rho)}.
\label{eq:singleton-alon-bound}
\end{equation}
\end{theorem}

\noindent\textbf{Proof sketch.}
Normalizing $C$ to unit diagonal yields the basic bound via a trace inequality and the stronger bound via Alon's rank theorem~\cite{alon2009perturbed}. 
Full proofs see Appendix~\ref{app:singleton-capacity}.

\noindent\textbf{Finding 3: Evaluation capacity limits singleton coverage.}
Even under oracle deletion, insufficient representation rank prevents target-specific linear scores from satisfying the selectivity criterion for all targets. 
For fixed $0<\rho<1/2$, Theorem~\ref{thm:singleton-capacity} requires this capacity to grow at least logarithmically with $n$. 
These bounds identify necessary capacity, leaving the coverage achieved by a particular evaluator to be measured. 
Establishing that coverage calls for a comprehensive controlled test under a common false-positive budget. 
To address arbitrary forget subsets, such a test must also distinguish complete from partial deletion, which motivates the following analysis for size-$K$ requests.

\subsection{Beyond Singleton Coverage}
\label{sec:subset-capacity}

\noindent\textbf{Selective subset coverage.}
For arbitrary subset requests, the evaluator must distinguish complete deletion from outcomes retaining any requested example.
Let $\mathcal S_K$ denote all size-$K$ subsets of $[n]$, where $K\leq n/2$. 
For  $S,T\in\mathcal S_K$, write $Z_T=Z_{\{u_i:i\in T\}}$ and use the same evaluation framework to assess request $S$ through $h_S(Z_T)$. 
Since the oracle deleting $T$ completes the request precisely when $T=S$, we then extend the selectivity criterion in Equation~\eqref{eq:shared-target-score} as
\begin{equation}
h_S(Z_S)\geq a,
\qquad
|h_S(Z_T)|\leq b
\quad (T\neq S).
\label{eq:subset-selectivity}
\end{equation}

\begin{corollary}[\textbf{Capacity for arbitrary subset coverage}]
\label{cor:subset-capacity}
Suppose a shared $d$-dimensional representation supports selective
scores for all requests in $\mathcal S_K$.
For $\binom{n}{K}^{-1/2}\leq\rho<1/2$,
a universal constant $c>0$ gives
\begin{equation}
d\geq c\,\frac{K\log(n/K)}{\rho^2\log(1/\rho)}.
\label{eq:subset-capacity-bound}
\end{equation}
\end{corollary}

\noindent\textbf{Proof sketch.}
Apply Theorem~\ref{thm:singleton-capacity} to the
$\binom{n}{K}$ subset requests and use
$\log\binom{n}{K}\geq K\log(n/K)$.
Details see Appendix~\ref{app:subset-capacity}. 

\noindent\textbf{Evidence for complete deletion.}
A score that distinguishes complete deletion of $S$ from deletions
that retain all its elements may still respond strongly when only
part of $S$ is deleted. 
For example, take $S=\{1,2\}$.
Deleting $\{3,4\}$ retains both requested examples, while deleting $\{1,3\}$ or $\{2,3\}$ retains one. 
Equation~\eqref{eq:subset-selectivity} requires all three scores
for $S$ to lie in $[-b,b]$. 
If $\psi(Z_S)$ is the midpoint of the representations of the two partial deletions, linearity makes $h_S(Z_S)$ the average of their scores, giving $h_S(Z_S)\leq b<a$. 
Thus, the following proposition extends this midpoint argument. 

\begin{proposition}[When controls preclude subset selectivity]
\label{prop:subset-control-selectivity}
Fix a request $S\in\mathcal S_K$ and let
$\mathcal C_S\subseteq\mathcal S_K\setminus\{S\}$
be the family of deletion sets used for comparison with $S$. 
The linear score constraints in Equation~\eqref{eq:subset-selectivity}
are infeasible for $S$ against $\mathcal C_S$ if and only if
there exist real coefficients $\lambda_T$ such
\begin{equation}
v_S=\sum_{T\in\mathcal C_S}\lambda_T v_T,
\qquad
b\sum_{T\in\mathcal C_S}|\lambda_T|<a.
\label{eq:subset-control-reconstruction}
\end{equation}
\end{proposition}

\noindent\textbf{Proof sketch.} 
Applying the comparison bounds to \eqref{eq:subset-control-reconstruction} gives $|w_S^\top v_S|\leq b\sum_T|\lambda_T|<a$.
Conversely, Farkas' lemma yields such a reconstruction when the score constraints in \eqref{eq:subset-selectivity} are infeasible. 
Details see Appendix~\ref{app:subset-controls}.

\noindent\textbf{Finding 4: Omitting overlapping deletions can
overstate subset coverage.}
Every disjoint deletion retains all members of $S$, so a score tracking deletion of a single member can pass all disjoint comparisons. 
A request covered by these comparisons loses selectivity precisely when adding overlapping deletions permits a reconstruction of $v_S$ satisfying \eqref{eq:subset-control-reconstruction}. 
Appending observables restores selectivity exactly when every such reconstruction fails to match the oracle representation for $S$ on at least one new coordinate.

\section{The CEO Framework}\label{sec:method}




In this section, we introduce \textbf{CEO} to assess unlearning coverage under limited observability. 
For each singleton request, our compositional benchmark of color--shape pairs directly measures forget and retain utility and provides a retraining oracle. 
We further define empirical target coverage as the fraction of targets meeting detection and false-positive criteria. 
Building on the capacity analysis, we expand the model observables to the higher-dimensional activation space of a frozen SAE trained on the original model.


\subsection{Measuring Singleton Coverage}
\label{sec:compositional-stress-test}

\begin{figure}[t]
    \centering
    \includegraphics[width=0.98\linewidth]{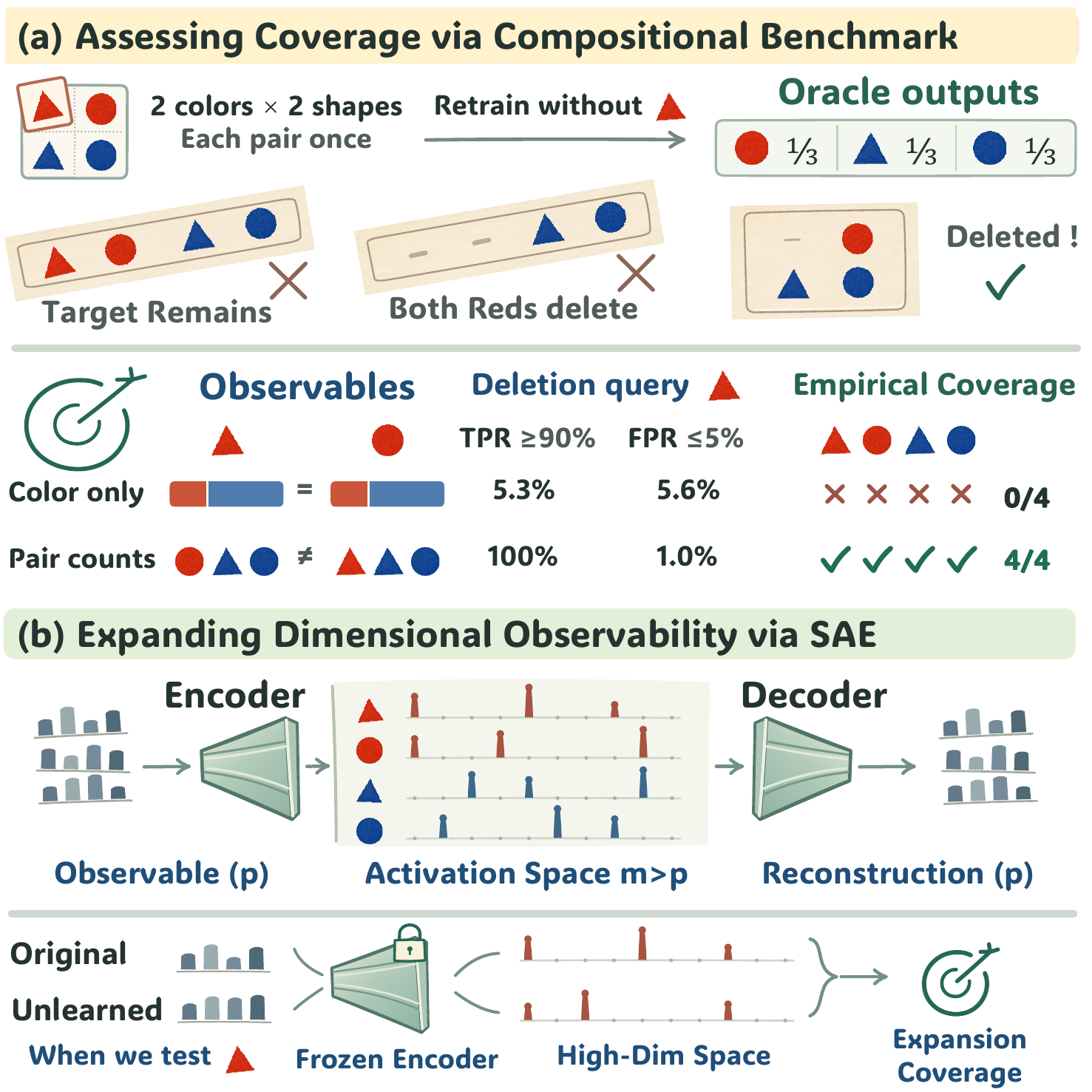}
    \vspace{-0.3cm}
\caption{\textbf{CEO evaluates coverage and narrow the observability gap.} Can unlearning get full coverage even on this simple task?}
    \label{fig:fig3}
    \vspace{-0.3cm}
\end{figure}

Retraining from scratch for every singleton request is costly, so we use a compositional task with a known reference distribution for each deletion to assess singleton coverage.  
We construct $\mathcal D$ by pairing each of $k$ colors with each of $k$ shapes exactly once, so removing any pair $u_i$ leaves a uniform empirical distribution on $\mathcal R_i=\mathcal D\setminus\{u_i\}$. 
A density model scores each pair from its color and shape indicators and normalizes the scores over all $n=k^2$ pairs, allowing us to measure the  remaining probability and the relative probabilities of the retained pairs after each edit. 
In Fig~\ref{fig:fig3}(a), omitting the red circle keeps the "red triangle" in training and leaves the same color counts as omitting the target itself. 
Using a single set of leave-one-out models as oracles and retention controls, we test whether unlearning methods can achieve full singleton coverage even in this simple task.


\subsection{Assessing the Coverage of Unlearning Methods}
\label{sec:measuring-request-coverage}

Under the singleton selectivity criterion in Eq.~\eqref{eq:target-selectivity}, the score for $u_i$ must be at least $a$ when $u_i$ is deleted and at most $b$ in magnitude when any other target is deleted.
To measure this selectivity empirically, we fix the sampling budget per trial and calibrate score thresholds on independent oracle runs.
On held-out trials, let $\widehat{\mathrm{TPR}}_i$ denote the rate at which the evaluator signals deletion of $u_i$ when $u_i$ is omitted, and $\widehat{\mathrm{FPR}}_{ji}$ the corresponding rate when $u_j$ is omitted and $u_i$ is retained.
We count $u_i$ as \textit{covered} if its detection rate is at least $\beta$ and its false-positive rate is at most $\alpha$ for every $j\ne i$.
Averaging over the targets gives  empirical coverage,
\begin{equation}
  \widehat{\operatorname{Cov}}_{\mathrm{obs}}
  = \mathbb{E}_i\!\bigl[
    \mathbf{1}\!\bigl\{\widehat{\mathrm{TPR}}_i \ge \beta,\,
    \max_{j\ne i}\widehat{\mathrm{FPR}}_{ji} \le \alpha\bigr\}
    \bigr].
  \label{eq:observable-request-coverage}
\end{equation}
For approximate unlearning methods, we label trials by the requested edit and report coverage alongside normalized residual target mass and retained-distribution error.

\begin{table*}[t]
\caption{\textbf{Coverage of singleton deletion requests on the color--shape benchmark.} 
A target is covered if its TPR is at least $90\%$ when its deletion is requested and its FPR is at most $5\%$ under every other request.
Values are percentages (mean $\pm$ SD over three training seeds).}
\label{tab:method_benchmark}
\centering
\begingroup
\definecolor{ceoTableInk}{HTML}{255C5F}
\definecolor{ceoTableRef}{HTML}{F3F5F4}
\definecolor{ceoTableSD}{HTML}{343D40}
\fontsize{8.10}{9.55}\selectfont
\renewcommand{\arraystretch}{1.00}
\setlength{\tabcolsep}{1.0pt}
\def\ceoSD{\normalfont\fontsize{6.85}{7.6}\selectfont\color{ceoTableSD}}
\definecolor{ceoRankFirst}{HTML}{0F3220}
\definecolor{ceoRankSecond}{HTML}{23462B}
\definecolor{ceoRankThird}{HTML}{35533B}
\def\ceoFirst#1{{\color{ceoRankFirst}\bfseries #1}}
\def\ceoSecond#1{{\color{ceoRankSecond}\itshape #1}}
\def\ceoThird#1{{\color{ceoRankThird}\underline{#1}}}
\definecolor{ceoFirstFill}{HTML}{8DC898}
\definecolor{ceoSecondFill}{HTML}{B9DFB8}
\definecolor{ceoThirdFill}{HTML}{DFF0D9}
\def\ceoFirstCell#1{\cellcolor{ceoFirstFill}\ceoFirst{#1}}
\def\ceoSecondCell#1{\cellcolor{ceoSecondFill}\ceoSecond{#1}}
\def\ceoThirdCell#1{\cellcolor{ceoThirdFill}\ceoThird{#1}}
\def\ceoKey#1#2{\begingroup\setlength{\fboxsep}{0pt}%
  \colorbox{#1}{\makebox[1.95em][c]{\strut #2}}\endgroup}
\def\ceoRowGap{\noalign{\vskip.45pt}}
\def\ceoHead#1{\multicolumn{1}{c}{\mbox{#1}}}
\resizebox{\linewidth}{!}{%
\begin{tabular}{@{}l*{3}{@{\hspace{5.0pt}}r@{\hspace{3.4pt}}r@{\hspace{3.4pt}}r}@{\hspace{1.0pt}}}
\toprule
 & \multicolumn{3}{c}{\textbf{16 targets},\enspace 60 samples}
 & \multicolumn{3}{c}{\textbf{64 targets},\enspace 252 samples}
 & \multicolumn{3}{c}{\textbf{256 targets},\enspace 1,020 samples}\\
\cmidrule(lr){2-4}\cmidrule(lr){5-7}\cmidrule(l){8-10}
\multicolumn{7}{@{}l}{\textcolor{ceoTableInk}{\bfseries A.}\enspace\textbf{Unlearning coverage}\quad A fixed joint-frequency evaluator scores all unlearning configurations.}
 & \multicolumn{3}{r@{}}{\fontsize{7.5}{8.5}\selectfont Column rank\enspace\ceoKey{ceoFirstFill}{\ceoFirst{1st}}\hspace{2pt}\ceoKey{ceoSecondFill}{\ceoSecond{2nd}}\hspace{2pt}\ceoKey{ceoThirdFill}{\ceoThird{3rd}}}\\[1.8pt]
Configuration
 & \ceoHead{Mean TPR\,\(\uparrow\)} & \ceoHead{Max.\ FPR\,\(\downarrow\)} & \ceoHead{\textcolor{ceoTableInk}{Coverage}\(\uparrow\)}
 & \ceoHead{Mean TPR\,\(\uparrow\)} & \ceoHead{Max.\ FPR\,\(\downarrow\)} & \ceoHead{\textcolor{ceoTableInk}{Coverage}\(\uparrow\)}
 & \ceoHead{Mean TPR\,\(\uparrow\)} & \ceoHead{Max.\ FPR\,\(\downarrow\)} & \ceoHead{\textcolor{ceoTableInk}{Coverage}\(\uparrow\)}\\
\midrule
Original & 2.18{\ceoSD\,\textpm\,0.06} & 3.58{\ceoSD\,\textpm\,\phantom{0}0.54} & 0.00{\ceoSD\,\textpm\,\phantom{0}0.00} & 1.90{\ceoSD\,\textpm\,\phantom{0}0.03} & 3.78{\ceoSD\,\textpm\,0.20} & 0.00{\ceoSD\,\textpm\,\phantom{0}0.00} & 1.84{\ceoSD\,\textpm\,0.02} & 3.97{\ceoSD\,\textpm\,0.20} & 0.00{\ceoSD\,\textpm\,0.00} \\\ceoRowGap
\noalign{\begingroup\color{ceoTableRef}%
  \hrule height\dimexpr\ht\strutbox+\dp\strutbox\relax
  \endgroup\vskip-\dimexpr\ht\strutbox+\dp\strutbox\relax}
Retraining & \ceoThirdCell{99.98}{\ceoSD\,\textpm\,0.00} & \ceoFirstCell{2.77}{\ceoSD\,\textpm\,\phantom{0}0.11} & \ceoFirstCell{100.00}{\ceoSD\,\textpm\,\phantom{0}0.00} & 99.95{\ceoSD\,\textpm\,\phantom{0}0.02} & \ceoFirstCell{3.26}{\ceoSD\,\textpm\,0.11} & \ceoFirstCell{100.00}{\ceoSD\,\textpm\,\phantom{0}0.00} & 99.90{\ceoSD\,\textpm\,0.01} & 4.04{\ceoSD\,\textpm\,0.20} & \ceoFirstCell{100.00}{\ceoSD\,\textpm\,0.00} \\
\addlinespace[1.6pt]
Finetune & 85.90{\ceoSD\,\textpm\,7.76} & \ceoThirdCell{3.09}{\ceoSD\,\textpm\,\phantom{0}0.46} & 52.08{\ceoSD\,\textpm\,43.90} & 39.35{\ceoSD\,\textpm\,14.36} & 4.62{\ceoSD\,\textpm\,0.23} & 0.00{\ceoSD\,\textpm\,\phantom{0}0.00} & 12.10{\ceoSD\,\textpm\,3.93} & 4.59{\ceoSD\,\textpm\,0.52} & 0.00{\ceoSD\,\textpm\,0.00} \\\ceoRowGap
GA & \ceoFirstCell{100.00}{\ceoSD\,\textpm\,0.00} & 100.00{\ceoSD\,\textpm\,\phantom{0}0.00} & 0.00{\ceoSD\,\textpm\,\phantom{0}0.00} & \ceoFirstCell{100.00}{\ceoSD\,\textpm\,\phantom{0}0.00} & 100.00{\ceoSD\,\textpm\,0.00} & 0.00{\ceoSD\,\textpm\,\phantom{0}0.00} & \ceoFirstCell{100.00}{\ceoSD\,\textpm\,0.00} & 100.00{\ceoSD\,\textpm\,0.00} & 0.00{\ceoSD\,\textpm\,0.00} \\\ceoRowGap
NPO & \ceoFirstCell{100.00}{\ceoSD\,\textpm\,0.00} & 100.00{\ceoSD\,\textpm\,\phantom{0}0.00} & 0.00{\ceoSD\,\textpm\,\phantom{0}0.00} & \ceoFirstCell{100.00}{\ceoSD\,\textpm\,\phantom{0}0.00} & 100.00{\ceoSD\,\textpm\,0.00} & 0.00{\ceoSD\,\textpm\,\phantom{0}0.00} & \ceoFirstCell{100.00}{\ceoSD\,\textpm\,0.00} & 100.00{\ceoSD\,\textpm\,0.00} & 0.00{\ceoSD\,\textpm\,0.00} \\\ceoRowGap
NegGrad+ & 99.96{\ceoSD\,\textpm\,0.05} & 6.74{\ceoSD\,\textpm\,\phantom{0}1.13} & 60.42{\ceoSD\,\textpm\,30.83} & \ceoFirstCell{100.00}{\ceoSD\,\textpm\,\phantom{0}0.00} & 17.55{\ceoSD\,\textpm\,6.07} & 2.08{\ceoSD\,\textpm\,\phantom{0}2.39} & \ceoFirstCell{100.00}{\ceoSD\,\textpm\,0.00} & 15.27{\ceoSD\,\textpm\,3.05} & \ceoSecondCell{16.54}{\ceoSD\,\textpm\,4.07} \\\ceoRowGap
NPO + retain & 99.92{\ceoSD\,\textpm\,0.05} & 4.79{\ceoSD\,\textpm\,\phantom{0}0.45} & \ceoSecondCell{97.92}{\ceoSD\,\textpm\,\phantom{0}3.61} & 99.98{\ceoSD\,\textpm\,\phantom{0}0.01} & 12.27{\ceoSD\,\textpm\,4.21} & \ceoThirdCell{21.35}{\ceoSD\,\textpm\,14.85} & \ceoFirstCell{100.00}{\ceoSD\,\textpm\,0.00} & 20.31{\ceoSD\,\textpm\,2.99} & 2.47{\ceoSD\,\textpm\,1.63} \\\ceoRowGap
NPO + cKL & \ceoThirdCell{99.98}{\ceoSD\,\textpm\,0.02} & 5.21{\ceoSD\,\textpm\,\phantom{0}0.94} & \ceoThirdCell{89.58}{\ceoSD\,\textpm\,13.01} & 99.98{\ceoSD\,\textpm\,\phantom{0}0.02} & 10.84{\ceoSD\,\textpm\,3.48} & \ceoSecondCell{29.17}{\ceoSD\,\textpm\,13.47} & \ceoFirstCell{100.00}{\ceoSD\,\textpm\,0.00} & 20.48{\ceoSD\,\textpm\,3.08} & \ceoThirdCell{3.26}{\ceoSD\,\textpm\,3.28} \\\ceoRowGap
SSD & 2.16{\ceoSD\,\textpm\,0.12} & 3.42{\ceoSD\,\textpm\,\phantom{0}0.10} & 0.00{\ceoSD\,\textpm\,\phantom{0}0.00} & 1.87{\ceoSD\,\textpm\,\phantom{0}0.06} & \ceoThirdCell{3.58}{\ceoSD\,\textpm\,0.06} & 0.00{\ceoSD\,\textpm\,\phantom{0}0.00} & 1.83{\ceoSD\,\textpm\,0.01} & 3.97{\ceoSD\,\textpm\,0.37} & 0.00{\ceoSD\,\textpm\,0.00} \\\ceoRowGap
Diag. Fisher & 63.10{\ceoSD\,\textpm\,7.75} & 70.51{\ceoSD\,\textpm\,19.75} & 0.00{\ceoSD\,\textpm\,\phantom{0}0.00} & 6.96{\ceoSD\,\textpm\,\phantom{0}0.14} & 8.37{\ceoSD\,\textpm\,0.60} & 0.00{\ceoSD\,\textpm\,\phantom{0}0.00} & 2.43{\ceoSD\,\textpm\,0.07} & 4.20{\ceoSD\,\textpm\,0.20} & 0.00{\ceoSD\,\textpm\,0.00} \\
\addlinespace[1.6pt]
SimNPO+R & 81.57{\ceoSD\,\textpm\,5.21} & 4.59{\ceoSD\,\textpm\,\phantom{0}1.03} & 18.75{\ceoSD\,\textpm\,16.54} & 37.09{\ceoSD\,\textpm\,10.77} & 6.54{\ceoSD\,\textpm\,0.35} & 0.00{\ceoSD\,\textpm\,\phantom{0}0.00} & 9.99{\ceoSD\,\textpm\,2.73} & 5.47{\ceoSD\,\textpm\,0.26} & 0.00{\ceoSD\,\textpm\,0.00} \\\ceoRowGap
WGA+R & 80.87{\ceoSD\,\textpm\,8.10} & 3.97{\ceoSD\,\textpm\,\phantom{0}1.05} & 20.83{\ceoSD\,\textpm\,18.04} & 33.47{\ceoSD\,\textpm\,12.46} & 5.14{\ceoSD\,\textpm\,0.30} & 0.00{\ceoSD\,\textpm\,\phantom{0}0.00} & 8.96{\ceoSD\,\textpm\,2.93} & 5.05{\ceoSD\,\textpm\,0.06} & 0.00{\ceoSD\,\textpm\,0.00} \\\ceoRowGap
UNDIAL & 80.44{\ceoSD\,\textpm\,7.58} & \ceoSecondCell{3.03}{\ceoSD\,\textpm\,\phantom{0}0.17} & 0.00{\ceoSD\,\textpm\,\phantom{0}0.00} & 38.05{\ceoSD\,\textpm\,13.19} & 4.72{\ceoSD\,\textpm\,1.00} & 0.00{\ceoSD\,\textpm\,\phantom{0}0.00} & 12.07{\ceoSD\,\textpm\,3.77} & 4.43{\ceoSD\,\textpm\,0.15} & 0.00{\ceoSD\,\textpm\,0.00} \\\ceoRowGap
RMU & 2.05{\ceoSD\,\textpm\,0.20} & 3.35{\ceoSD\,\textpm\,\phantom{0}0.06} & 0.00{\ceoSD\,\textpm\,\phantom{0}0.00} & 1.97{\ceoSD\,\textpm\,\phantom{0}0.15} & 3.61{\ceoSD\,\textpm\,0.20} & 0.00{\ceoSD\,\textpm\,\phantom{0}0.00} & 2.05{\ceoSD\,\textpm\,0.12} & \ceoThirdCell{3.84}{\ceoSD\,\textpm\,0.06} & 0.00{\ceoSD\,\textpm\,0.00} \\\ceoRowGap
NegMerge & 1.48{\ceoSD\,\textpm\,0.89} & 5.31{\ceoSD\,\textpm\,\phantom{0}2.76} & 0.00{\ceoSD\,\textpm\,\phantom{0}0.00} & 1.90{\ceoSD\,\textpm\,\phantom{0}0.05} & 5.89{\ceoSD\,\textpm\,1.64} & 0.00{\ceoSD\,\textpm\,\phantom{0}0.00} & 1.79{\ceoSD\,\textpm\,0.28} & 4.72{\ceoSD\,\textpm\,1.07} & 0.00{\ceoSD\,\textpm\,0.00} \\\ceoRowGap
Saliency-NegGrad+ & 19.85{\ceoSD\,\textpm\,1.09} & 10.42{\ceoSD\,\textpm\,\phantom{0}1.07} & 0.00{\ceoSD\,\textpm\,\phantom{0}0.00} & 22.29{\ceoSD\,\textpm\,\phantom{0}1.12} & 15.62{\ceoSD\,\textpm\,0.93} & 0.00{\ceoSD\,\textpm\,\phantom{0}0.00} & 26.65{\ceoSD\,\textpm\,1.02} & 25.07{\ceoSD\,\textpm\,0.73} & 0.00{\ceoSD\,\textpm\,0.00} \\
\addlinespace[1.6pt]
Amnesiac rollback & 2.19{\ceoSD\,\textpm\,0.19} & 3.55{\ceoSD\,\textpm\,\phantom{0}0.15} & 0.00{\ceoSD\,\textpm\,\phantom{0}0.00} & 1.92{\ceoSD\,\textpm\,\phantom{0}0.07} & 3.91{\ceoSD\,\textpm\,0.20} & 0.00{\ceoSD\,\textpm\,\phantom{0}0.00} & 1.86{\ceoSD\,\textpm\,0.03} & 4.56{\ceoSD\,\textpm\,0.37} & 0.00{\ceoSD\,\textpm\,0.00} \\\ceoRowGap
Amnesiac+repair & 77.84{\ceoSD\,\textpm\,7.52} & 3.45{\ceoSD\,\textpm\,\phantom{0}0.76} & 10.42{\ceoSD\,\textpm\,13.01} & 19.93{\ceoSD\,\textpm\,\phantom{0}4.80} & 4.56{\ceoSD\,\textpm\,0.11} & 0.00{\ceoSD\,\textpm\,\phantom{0}0.00} & 3.23{\ceoSD\,\textpm\,0.10} & 4.13{\ceoSD\,\textpm\,0.11} & 0.00{\ceoSD\,\textpm\,0.00} \\\ceoRowGap
BlindSpot-event & 2.01{\ceoSD\,\textpm\,0.02} & 3.42{\ceoSD\,\textpm\,\phantom{0}0.10} & 0.00{\ceoSD\,\textpm\,\phantom{0}0.00} & 1.87{\ceoSD\,\textpm\,\phantom{0}0.02} & 3.65{\ceoSD\,\textpm\,0.20} & 0.00{\ceoSD\,\textpm\,\phantom{0}0.00} & 1.86{\ceoSD\,\textpm\,0.01} & \ceoFirstCell{3.81}{\ceoSD\,\textpm\,0.10} & 0.00{\ceoSD\,\textpm\,0.00} \\\ceoRowGap
FullTeacher & 1.95{\ceoSD\,\textpm\,0.14} & 3.32{\ceoSD\,\textpm\,\phantom{0}0.10} & 0.00{\ceoSD\,\textpm\,\phantom{0}0.00} & 1.87{\ceoSD\,\textpm\,\phantom{0}0.04} & \ceoSecondCell{3.52}{\ceoSD\,\textpm\,0.17} & 0.00{\ceoSD\,\textpm\,\phantom{0}0.00} & 1.82{\ceoSD\,\textpm\,0.03} & \ceoFirstCell{3.81}{\ceoSD\,\textpm\,0.10} & 0.00{\ceoSD\,\textpm\,0.00} \\
\midrule
\multicolumn{10}{@{}l@{}}{\textcolor{ceoTableInk}{\bfseries B.}\enspace\textbf{Oracle observability}\quad Six evaluators score the same output samples from leave-one-out retraining.}\\[1.8pt]
Evaluator
 & \ceoHead{Mean TPR\,\(\uparrow\)} & \ceoHead{Max.\ FPR\,\(\downarrow\)} & \ceoHead{\textcolor{ceoTableInk}{Coverage}\(\uparrow\)}
 & \ceoHead{Mean TPR\,\(\uparrow\)} & \ceoHead{Max.\ FPR\,\(\downarrow\)} & \ceoHead{\textcolor{ceoTableInk}{Coverage}\(\uparrow\)}
 & \ceoHead{Mean TPR\,\(\uparrow\)} & \ceoHead{Max.\ FPR\,\(\downarrow\)} & \ceoHead{\textcolor{ceoTableInk}{Coverage}\(\uparrow\)}\\
\midrule
\multicolumn{10}{@{}l@{}}{\textcolor{ceoTableInk}{\itshape Separate counts}\enspace One count per color and one per shape, discarding their pairings.}\\
Marginal Linear & 11.66{\ceoSD\,\textpm\,0.37} & 4.62{\ceoSD\,\textpm\,\phantom{0}0.15} & 0.00{\ceoSD\,\textpm\,\phantom{0}0.00} & 9.72{\ceoSD\,\textpm\,\phantom{0}0.17} & 6.77{\ceoSD\,\textpm\,0.79} & 0.00{\ceoSD\,\textpm\,\phantom{0}0.00} & 7.78{\ceoSD\,\textpm\,0.07} & 7.68{\ceoSD\,\textpm\,0.65} & 0.00{\ceoSD\,\textpm\,0.00} \\
Marginal Quadratic & 16.15{\ceoSD\,\textpm\,0.23} & 6.35{\ceoSD\,\textpm\,\phantom{0}0.20} & 0.00{\ceoSD\,\textpm\,\phantom{0}0.00} & 10.82{\ceoSD\,\textpm\,\phantom{0}0.14} & 7.62{\ceoSD\,\textpm\,1.13} & 0.00{\ceoSD\,\textpm\,\phantom{0}0.00} & 9.15{\ceoSD\,\textpm\,0.08} & 8.85{\ceoSD\,\textpm\,0.60} & 0.00{\ceoSD\,\textpm\,0.00} \\
Marginal Likelihood & 17.88{\ceoSD\,\textpm\,0.09} & 6.28{\ceoSD\,\textpm\,\phantom{0}0.11} & 0.00{\ceoSD\,\textpm\,\phantom{0}0.00} & 11.52{\ceoSD\,\textpm\,\phantom{0}0.14} & 7.94{\ceoSD\,\textpm\,0.94} & 0.00{\ceoSD\,\textpm\,\phantom{0}0.00} & 9.26{\ceoSD\,\textpm\,0.06} & 9.05{\ceoSD\,\textpm\,0.90} & 0.00{\ceoSD\,\textpm\,0.00} \\
Redundant Linear & 11.66{\ceoSD\,\textpm\,0.37} & 4.62{\ceoSD\,\textpm\,\phantom{0}0.15} & 0.00{\ceoSD\,\textpm\,\phantom{0}0.00} & 9.72{\ceoSD\,\textpm\,\phantom{0}0.17} & 6.77{\ceoSD\,\textpm\,0.79} & 0.00{\ceoSD\,\textpm\,\phantom{0}0.00} & 7.78{\ceoSD\,\textpm\,0.07} & 7.68{\ceoSD\,\textpm\,0.65} & 0.00{\ceoSD\,\textpm\,0.00} \\
\addlinespace[1.3pt]
\multicolumn{10}{@{}l@{}}{\textcolor{ceoTableInk}{\itshape Joint counts}\enspace One count per color–shape pair.}\\
Joint Frequency & 99.98{\ceoSD\,\textpm\,0.00} & 2.77{\ceoSD\,\textpm\,\phantom{0}0.11} & {\bfseries\color{ceoTableInk}100.00}{\ceoSD\,\textpm\,\phantom{0}0.00} & 99.95{\ceoSD\,\textpm\,\phantom{0}0.02} & 3.26{\ceoSD\,\textpm\,0.11} & {\bfseries\color{ceoTableInk}100.00}{\ceoSD\,\textpm\,\phantom{0}0.00} & 99.90{\ceoSD\,\textpm\,0.01} & 4.04{\ceoSD\,\textpm\,0.20} & {\bfseries\color{ceoTableInk}100.00}{\ceoSD\,\textpm\,0.00} \\
Zero Count & 99.98{\ceoSD\,\textpm\,0.00} & 2.77{\ceoSD\,\textpm\,\phantom{0}0.11} & {\bfseries\color{ceoTableInk}100.00}{\ceoSD\,\textpm\,\phantom{0}0.00} & 99.95{\ceoSD\,\textpm\,\phantom{0}0.02} & 3.26{\ceoSD\,\textpm\,0.11} & {\bfseries\color{ceoTableInk}100.00}{\ceoSD\,\textpm\,\phantom{0}0.00} & 99.90{\ceoSD\,\textpm\,0.01} & 4.04{\ceoSD\,\textpm\,0.20} & {\bfseries\color{ceoTableInk}100.00}{\ceoSD\,\textpm\,0.00} \\
\bottomrule
\end{tabular}}%
\endgroup
\vspace{-0.5cm}
\end{table*}

\subsection{Dimensional Expansion for Observability}
\label{sec:expanding-observability}

Motivated by the rank requirement on the shared representation in Theorem~\ref{thm:singleton-capacity}, we introduce \emph{Dimensional Expansion} to learn a family of nonlinear maps for individual model responses. 
Given an observable $z_M(x)\in\mathbb{R}^p$, each map $\Phi:\mathbb{R}^p\to\mathbb{R}^m$ with $m>p$ yields an expanded response $\widetilde z_M(x)=\Phi(z_M(x))$. 
Applying $\Phi$ before pooling lets us measure the average activation of features that depend jointly on several coordinates of an individual response. 
To compare these activations before and after deletion, we learn each map from embeddings of the original model $M_{\mathcal D}$ and keep it fixed across deletion requests.

We learn $\Phi$ with an overcomplete sparse autoencoder (SAE), which encodes each $p$-dimensional embedding as a sparse pattern of activations over a shared dictionary of $m>p$ learned features~\cite{luo2026vocabulary}. 
At each expansion width, we train SAE to reconstruct embeddings of the original model $M_{\mathcal D}$. 
Given an observable $z_M(x)$, the SAE encoder retains the $\kappa$ largest activations of learned affine map $A_{\mathrm e}$, yielding
\begin{equation}
  h_M(x)
  = \operatorname{TopK}_{\kappa}\!\bigl([A_{\mathrm e}(z_M(x))]_+\bigr)
  \in \mathbb{R}^{m}.
  \label{eq:sae-expansion-training}
\end{equation}
We freeze the encoder and normalize $h_M(x)$ using original-model statistics to obtain the expanded observable $\widetilde z_M(x)=\Phi(z_M(x))$ for every deletion request.
For a queried image $u_i$, we use the root-mean-square difference between its expanded feature activations under the original model $M_{\mathcal D}$ and the post-deletion model $M$ as the detection score
\begin{equation}
  s_i^\Phi(M)
  = \frac{\bigl\|\widetilde z_M(u_i)
           - \widetilde z_{M_{\mathcal D}}(u_i)\bigr\|_2}{\sqrt{m}}.
  \label{eq:sparse-observable-expansion}
\end{equation}
We apply the same scoring rule to raw embeddings and calibrate thresholds separately for each representation on independent oracle runs.
With the model pairs, observation budget, and target-retaining controls fixed, we compare the per-target decisions in Eq.~\eqref{eq:observable-request-coverage} across repeated trials to identify which targets gain or lose coverage after expansion.

\section{Experiments}\label{sec:exp}
To evaluate CEO and existing unlearning methods, we ask five research questions. 
\textbf{RQ1 Unlearning coverage.} Can current methods forget any color--shape pair and preserve utility on the retained?
\textbf{RQ2 Oracle observability.} Which known deletions do different observables reveal at a fixed budget?
\textbf{RQ3 Dimensional Expansion.} When does expansion reveal missed deletions in the same model responses?
\textbf{RQ4 Empirical implications.} Do rank and partial-deletion controls affect coverage as predicted?
\textbf{RQ5 CEO Findings.} What can we learn about the conclusions from the results?

\begin{figure*}[!t]
    \centering
    \includegraphics[width=1\linewidth]{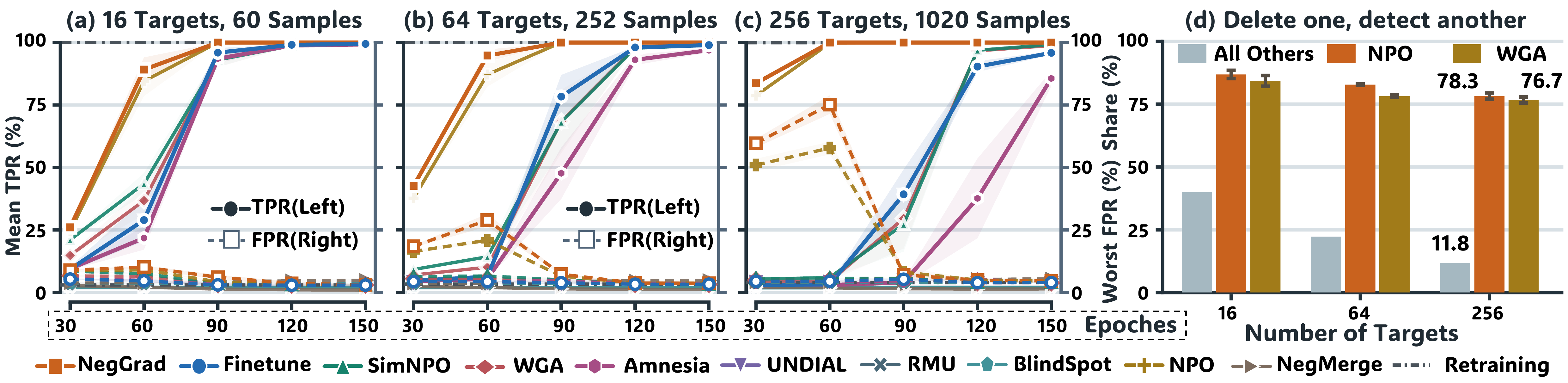}
\caption{\textbf{Detecting color--shape deletion requests.}
(a--c) Mean TPR (solid) and worst-case FPR (dashed) across training steps.
(d) At 60 updates and 256 targets, pairs sharing the target's color or shape make up 11.8\% of other pairs and over 75\% of false alarms, respectively.}
    \vspace{-0.1cm}
    \label{fig:fig4}
    \vspace{-0.3cm}
\end{figure*}

\begin{figure}
    \centering
    \includegraphics[width=1\linewidth]{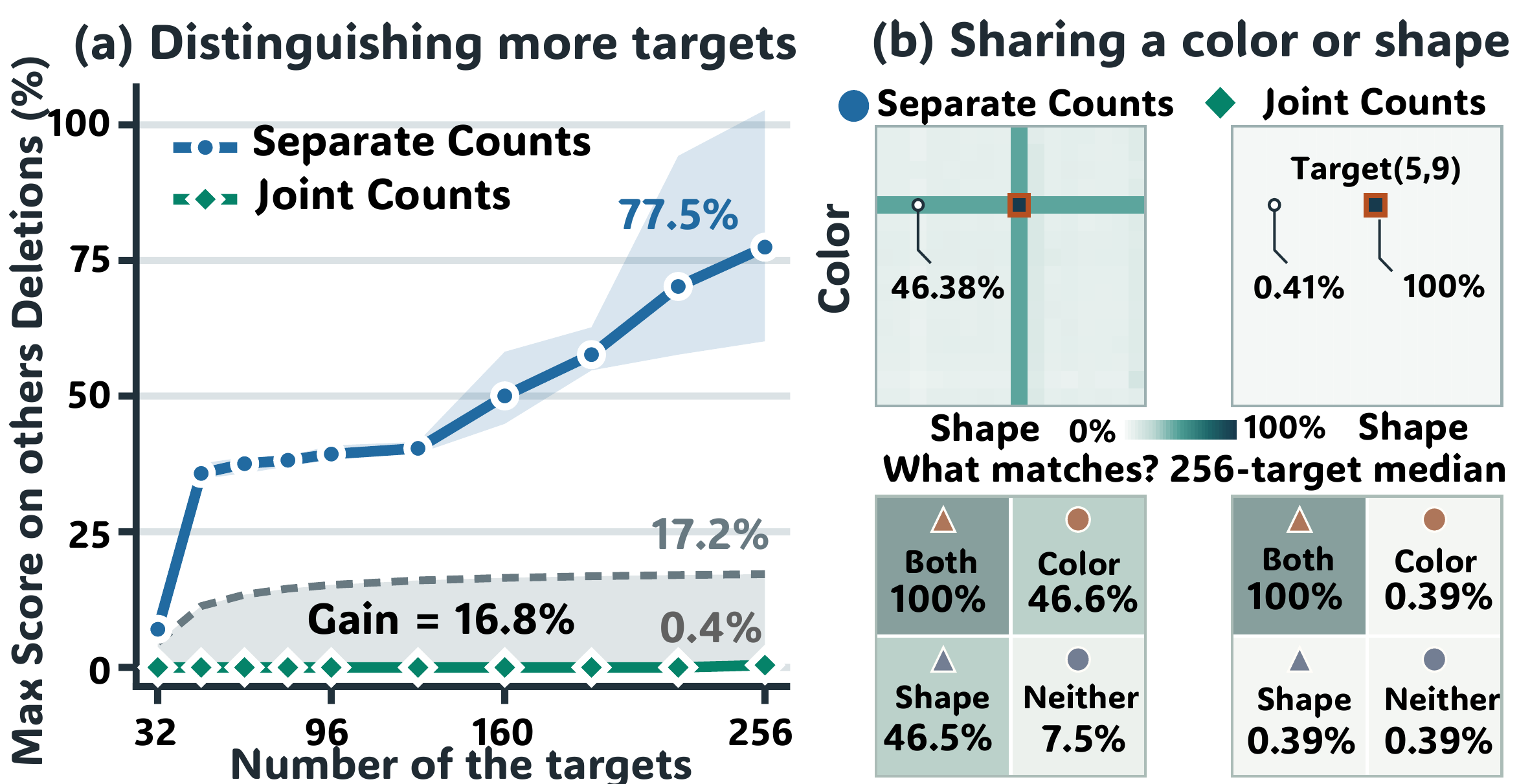}
\caption{\textbf{Non-target scores relative to the target deletion.}
(a) Maximum scores and the rank-based lower bound for separate counts (gray). (b) Shared color raises scores under separate counts.}
    \vspace{-0.2cm}
    \label{fig:fig5}
    \vspace{-0.3cm}
\end{figure}

\subsection{Experimental Setup}
\label{sec:experiment_setup}

\noindent\textbf{Benchmarks.}
For vision models, we use a synthetic color--shape task and two real-world image datasets, \mbox{Aircraft}~\cite{maji2013aircraft} and \mbox{Pets}~\cite{parkhi2012pets}. 
For language models, we construct two synthetic tasks, one pairing people with passcodes and the other pairing researchers with research stations. 
Each synthetic task has variants with 16, 64, and 256 target records. At each size, we repeatedly sample deletion requests targeting either a single record or a subset of multiple records. 
We evaluate ResNet-18~\cite{he2016resnet} and DINOv2-small~\cite{oquab2024dinov2}. 
For language models, we adapt \mbox{GPT-2}~\cite{radford2019gpt2} and
\mbox{Qwen2.5-3B}~\cite{qwen2024qwen25} with low-rank adaptation
(LoRA)~\cite{hu2022lora} on 1600 fictional facts, using 50 as deletion targets. Details see Appendix~\ref{app:general_setup}.

\noindent \textbf{Unlearning methods.}
We score the 20 color--shape configurations in Table~\ref{tab:method_benchmark}A with the same joint-frequency evaluator, using the original model as the no-edit control and leave-one-out retraining as the deletion oracle. 
The baselines span retain-set fine-tuning, gradient and preference-based updates~\cite{kurmanji2023unbounded,zhang2024npo}, distillation~\cite{chundawat2023badteaching}, and edits to weights or representations~\cite{foster2024ssd,li2024wmdp}.
Details see Appendices~\ref{app:general_setup}.

\noindent \textbf{Implementation details.}
After calibration on independent oracle runs, we count a target as covered if its TPR is at least $90\%$ and its FPR is at most $5\%$ for each other deletion. 
For $n$ color--shape targets, evaluators share the same $B=4(n-1)$ output samples per trial by default, and we report means and standard deviations over three seeds. 
For approximate edits, coverage measures recognition of the request, so we also report normalized residual target mass and the total variation distance of the retained-pair distribution from uniform. 
For language models, we report forget and retain accuracy and the mean absolute gap in target negative log-likelihood between each edited model and its oracle.

\begin{figure*}[t]
    \centering
    \includegraphics[width=1\linewidth]{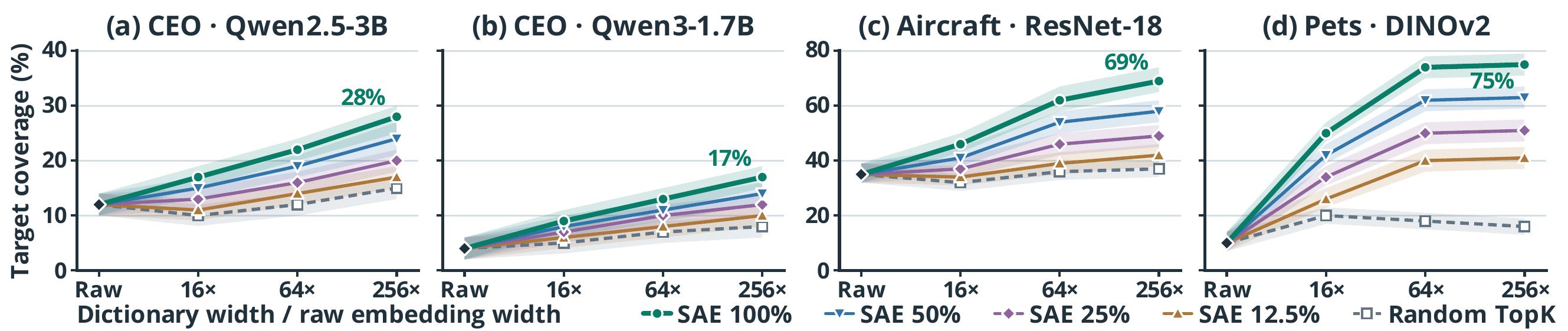}
\caption{\textbf{Target coverage after dimensional expansion.}
We compare SAEs trained on 12.5\%--100\% of the data at matched widths.}
    \vspace{-0.3cm}
    \label{fig:fig6}
    \vspace{-0.3cm}
\end{figure*}

\begin{findingsbox}{Findings for RQ1 (Full Coverage Is Rare)}
\noindent\textbf{Finding 1:} Even on this simple toy task, few approximate unlearning settings cover all requests at a fixed sampling budget.

\noindent\textbf{Finding 2:} With more training steps, several methods achieve substantially higher target coverage under this stress test.

\noindent\textbf{Finding 3:} Coverage gaps stem mainly from poor retention.
\end{findingsbox}

\vspace{-0.3cm}
\subsection{Evaluating Unlearning Coverage (RQ1)}
\label{sec:method_diagnosis}

Table~\ref{tab:method_benchmark}A compares 20 configurations on every singleton color--shape request using the same joint-frequency evaluator. 
At 16, 64, and 256 pairs, retraining covers every target, while the best mean request coverage among the approximate edits shown falls from $97.92\%$ to $29.17\%$ and then $16.54\%$. 
Across training budgets, five of 28 configurations cover all 256 targets, each after 800 updates. 
At 256 pairs, GA and NPO nearly eliminate the requested pair's probability but distort the retained distribution ($\approx 12\%$). Both reach $100\%$ mean TPR and $100\%$ worst FPR, yielding $0\%$ coverage. 
In Fig.~\ref{fig:fig4}(d), pairs sharing the target's color or shape make up only $11.8\%$ of alternatives yet account for $78.3\%$ of WGA FPR and $76.7\%$ of NPO FPR after 60 updates.

\begin{table}[t]
  \centering
  \caption{\textbf{How partial deletions limit subset selectivity?} We minimize the largest absolute score and count requests at a 25\% limit.}
  \label{tab:rq4_partial_controls}
  \fontsize{8.8}{10.0}\selectfont
  \setlength{\tabcolsep}{0pt}
  \setlength{\aboverulesep}{1.6pt}
  \setlength{\belowrulesep}{1.8pt}
  \renewcommand{\arraystretch}{1.08}
  \begin{tabular*}{\linewidth}{@{\extracolsep{\fill}}lrccc@{}}
    \toprule
    & & \multicolumn{2}{c}{Worst control score (\%)} & Requests\\
    \cmidrule(lr){3-4}
    Controls tested & Count & Uniform & Retrained & within limit\\
    \midrule
    \multicolumn{5}{@{}l}{\textbf{Pairs} \quad For $\{1,2\}$, $\{3,4\}$ is disjoint and $\{1,3\}$ is partial.}\\
    Disjoint only     &  91 & 14.29 & 14.30--17.54 & 120/120\\
    With 28 partial   &  91 & 42.86 & 42.04--44.00 & \textbf{0}/120\\
    All controls      & 119 & 42.86 & 42.26--44.00 & \textbf{0/120}\\
    \midrule
    \multicolumn{5}{@{}l}{\textbf{Triples} \quad For $\{1,2,3\}$, $\{1,2,4\}$ deletes two requested records.}\\
    Disjoint only     & 286 & 23.08 & 14.30--17.54 & 560/560\\
    Up to one shared & 520 & 23.08 & 56.62--67.54 & 560/560\\
    Up to two shared & 559 & 58.97 & 72.90--84.35 & \textbf{0}/560\\
    \bottomrule
  \end{tabular*}
  \vspace{-0.3cm}
\end{table}
\begin{figure}[t]
    \centering
    \includegraphics[width=1\linewidth]{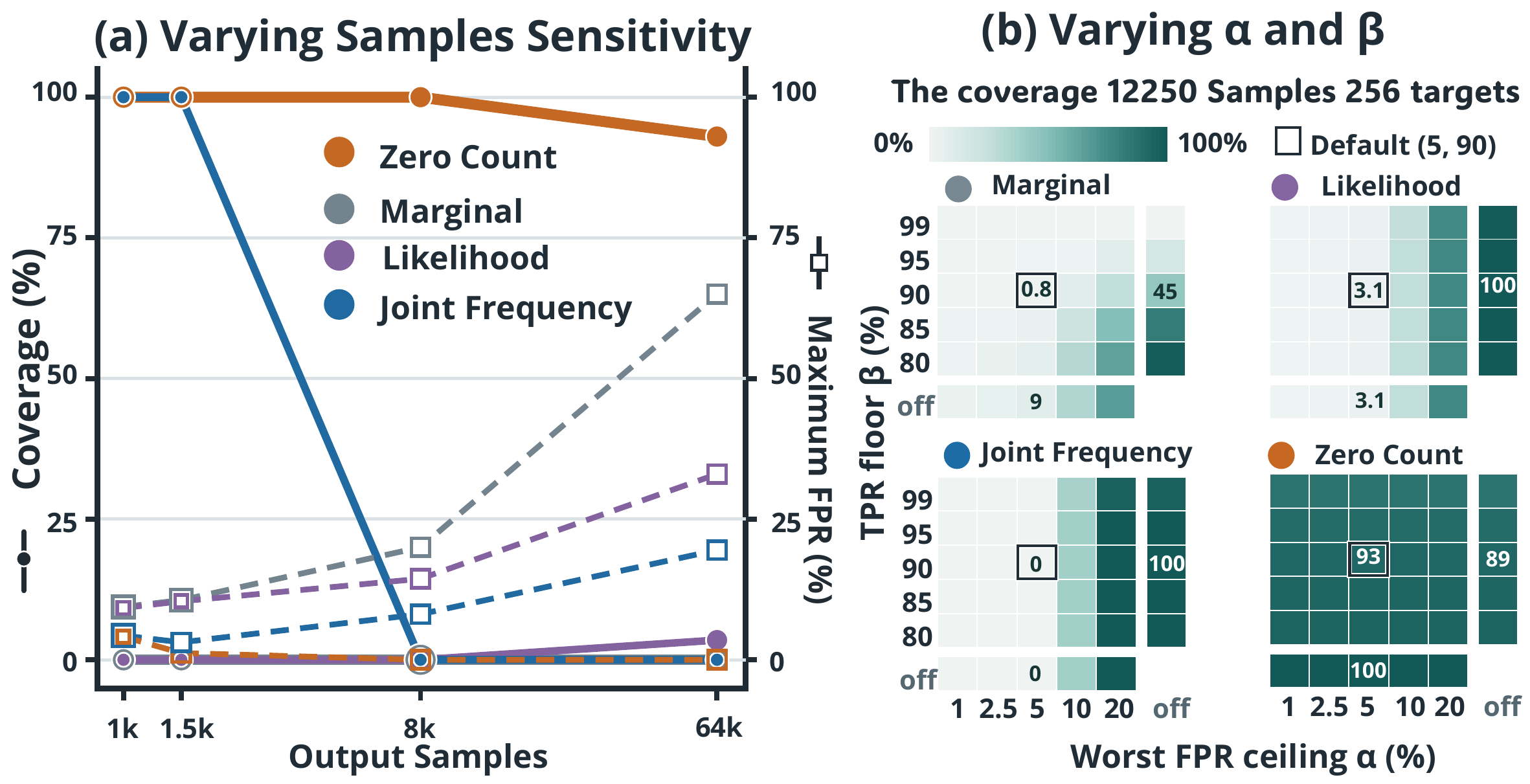}
\caption{\textbf{Coverage across budgets and thresholds.}
(a) Zero Count covers all targets, Joint Frequency covers none at 8192 samples. 
(b) Boxes mark default 5\% FPR ceiling and 90\% TPR floor.}
    \vspace{-0.2cm}
    \label{fig:fig7}
    \vspace{-0.3cm}
\end{figure}

\vspace{-0.1cm}
\subsection{Detecting Known Deletions (RQ2)}

To isolate the evaluator's role, we compare six evaluators on the same output draws from each leave-one-out model used in RQ1 (Table~\ref{tab:method_benchmark}B).
Across all three task sizes, none of the four evaluators that count colors and shapes separately covers a single target.
Joint Frequency and Zero Count both cover every target, including all 256 pairs with 1,020 draws per trial.
At this size, preserving the pairings reduces the largest score elicited by a different deletion from 77.5\% to 0.4\% of the score for deleting the queried pair (Fig.~\ref{fig:fig5}(a)).
When the deleted pair shares only the queried pair's color or shape, the median score on this scale falls from about 46.5\% with separate counts to 0.39\% with joint counts (Fig.~\ref{fig:fig5}(b)).

\begin{findingsbox}{Findings for RQ2 (Blind Spots in Separate Counts)}
\noindent\textbf{Finding.} Separate color and shape counts cover no targets.
Counting pairs in those same oracle outputs covers them all.
\end{findingsbox}

\subsection{Recovering Missed Deletions (RQ3)}
We revisit the 256 targets missed by separate counts in RQ2 using the same 1,020 oracle outputs per trial. 
We keep the decoder and 288-feature width fixed and vary only when the evaluator forms color--shape interactions. 
Expanding the pooled counts leaves every target uncovered. Expanding each output before pooling preserves the observed pairings and recovers all 256.
We then evaluate the same language and vision embeddings in the higher-dimensional activation space of the frozen SAE trained on the original model. 
Across all four settings, the full-data SAE at $256\times$ width covers more targets than either raw embeddings or a Random TopK encoder of matched width and sparsity (Fig.~\ref{fig:fig6}).

\begin{findingsbox}{Findings for RQ3 (Recoverable Blind Spots)}
\noindent\textbf{Finding.} Dimensional Expansion via SAE can reveal previously missed deletions by mapping the same low-dimensional observables into a higher-dimensional activation space.
\end{findingsbox}

\begin{findingsbox}{Findings for RQ4 (Distinguishing Deletions)}
\noindent\textbf{Finding 1.} For linear scores, the rank of the shared representation bounds singleton coverage at fixed selectivity.

\noindent\textbf{Finding 2.} A score tracking one member can pass disjoint controls without verifying deletion of the whole set.
\end{findingsbox}

\subsection{Capacity Limits and Partial Deletions (RQ4)}
We normalize each request's linear score to one on its own oracle deletion and measure its largest absolute response to a control deletion.
For 256 singleton targets, this maximum averages $77.5\%$ across seeds with separate color and shape counts (centered rank~30). Joint counts reduce it to $0.4\%$ on the same outputs, below the $17.2\%$ floor that Theorem~\ref{thm:singleton-capacity} imposes at rank~30 (Fig.~\ref{fig:fig5}(a)). 
For subset requests, the same criterion must hold even when a control deletes all but one requested member. 
On the uniform oracle, replacing 28 of 91 disjoint controls with such partial deletions raises the best attainable worst-case score for pairs from $14.29\%$ to $42.86\%$ (Table~\ref{tab:rq4_partial_controls}). 
For all 120 pairs, the mixed controls reconstruct the complete-deletion vector with $\sum_T|\lambda_T|<4$, so Proposition~3 rules out the $25\%$ cap.

\begin{table}[t]
  \centering
  \captionsetup{position=top,skip=3pt,font=small,
                justification=raggedright,singlelinecheck=false}
  \caption{\textbf{Coverage falls from all 256 targets to none.}
    Rates (\%) compare expansion \emph{before} \(\to\) \emph{after}
    pooling with a fixed map and decoder ($B=1{,}020$).
    FPR averages over non-target deletions.}
  \label{tab:pooling_controls}
  \fontsize{9.5}{10.3}\selectfont
  \setlength{\tabcolsep}{3.2pt}
  \setlength{\aboverulesep}{1.7pt}
  \setlength{\belowrulesep}{1.7pt}
  \setlength{\fboxsep}{1pt}
  \providecolor{coverblue}{RGB}{49,103,148}
  \newcommand*{\poolpair}[2]{#1\,\ensuremath{\to}\,#2}
  \newcommand*{\poolcov}[2]{%
    \colorbox{coverblue!12}{\color{coverblue!60!black}\bfseries\poolpair{#1}{#2}}}
  \resizebox{\linewidth}{!}{%
    \begin{tabular}{@{}lcccc@{\hspace{2.5pt}}}
      \toprule
      Features & Width & \textbf{Coverage}\,\(\uparrow\)
        & TPR\,\(\uparrow\) & Mean FPR\,\(\downarrow\) \\
      \midrule
      ReLU        & 288 & \poolcov{100}{0} & \poolpair{99.89}{4.56} & \poolpair{1.82}{4.26} \\
      Interaction & 288 & \poolcov{100}{0} & \poolpair{99.89}{8.97} & \poolpair{1.82}{2.47} \\
      \midrule[0.3pt]
      Raw         &  32 &   0 &  8.17 & 2.19 \\
      Linear      & 288 &   0 &  8.17 & 2.19 \\
      Joint counts & 256 & 100 & 99.89 & 1.82 \\
      \bottomrule
    \end{tabular}%
  }%
  \vspace{-0.5cm}
\end{table}

\subsection{When Decision Rules Limit Coverage (RQ5)}
RQ2 showed that color--shape pair counts can support full coverage. 
We apply Joint Frequency and Zero Count to identical pair counts at each sampling budget (Fig.~\ref{fig:fig7}(a)). 
With 1,020 draws, both rules cover all 256 targets.
With 8,192 draws, Zero Count still covers all 256, while the Joint Frequency rule reaches $100\%$ mean TPR and covers none. 
In the threshold sweep (Fig.~\ref{fig:fig7}(b)), every target meets the $90\%$ TPR floor for Joint Frequency, but its FPR exceeds $5\%$ under at least one deletion that retains it.

\begin{findingsbox}{Findings for RQ5 (Rule Dependent Blind Spots)}
\noindent\textbf{Finding.} Even when support full coverage, the false alarms of the evaluation rule can leave every target uncovered.
\end{findingsbox}

\vspace{-0.3cm}
\section{Conclusion}\label{sec:conc}
\vspace{-0.1cm}
In this work, we ask which deletion requests an unlearning method can fulfill while preserving utility, a question that evaluations on selected  sets often leave open. 
Full coverage was rare for approximate methods on our benchmark, and retraining oracles showed that even exact deletions could go undetected.  Guided by our rank bounds for linear scores, CEO uses a frozen SAE to reveal deletions missed in raw embeddings. Since retraining is costly, we aim to estimate coverage for larger subsets with fewer oracle runs.

\section*{LLM Usage Statement}
Generative AI tools (ChatGPT and Gemini) were used to assist with baseline reproduction, refining grammar and reviewing mathematical proofs. 
The authors take full responsibility for the final manuscript and experimental results.
\bibliographystyle{ieeenat_fullname}
\bibliography{main}

\clearpage
\appendix
\section{Experimental Details}
\label{app:experimental_details}
This appendix first describes the datasets, implementations, baselines,
and evaluation protocol shared across experiments
(Section~\ref{app:general_setup}). It then gives the controlled comparisons
and experiment-specific settings for each main-text table and figure
(Section~\ref{app:display_settings}).

\subsection{General setup and implementations}
\label{app:general_setup}
\subsubsection{Notation}
\label{app:notation}
Table~\ref{tab:app_notation} summarizes the notation used in the experiments.
The number of candidate targets $n$ need not equal the full training-set
size. Method-specific temperatures use subscripts, such as
$\beta_{\mathrm{NPO}}$, to distinguish them from the detection floor $\beta$.

\begin{table*}[t]
\centering
\caption{Notation for the datasets, model observations, and evaluation protocol.}
\label{tab:app_notation}
\small
\setlength{\tabcolsep}{5pt}
\renewcommand{\arraystretch}{1.05}
\begin{tabular}{@{}p{.22\textwidth}p{.74\textwidth}@{}}
\toprule
\textbf{Symbol} & \textbf{Definition} \\
\midrule
\multicolumn{2}{@{}l@{}}{\textit{Data and models}} \\
$\mathcal D,\mathcal F,\mathcal R$ & Training set, requested forget set, and retain set $\mathcal R=\mathcal D\setminus\mathcal F$. \\
$\mathcal A,\mathcal U$ & Training and unlearning algorithms. \\
$M_{\mathcal D}$ & Original model $\mathcal A(\mathcal D)$. \\
$M_{\mathcal R},M^u$ & Retraining oracle $\mathcal A(\mathcal R)$ and edited model $\mathcal U(M_{\mathcal D},\mathcal F)$. \\
$M$ & Model currently being evaluated. \\
$u_i,n$ & Queried training record and number of candidate deletion targets. \\
$k$ & Number of colors and shapes in each dictionary; $n=k^2$ in the compositional task. \\
$S,T,\mathcal S_K$ & Requested/deleted index subsets and the family of size-$K$ subsets. \\
$\mathcal C_S$ & Control-deletion family for request $S$. \\
\midrule
\multicolumn{2}{@{}l@{}}{\textit{Observations and representations}} \\
$\mathcal Q$ & Fixed query inputs used to collect model observations. \\
$z_M(x),p$ & Per-input model observable and its raw dimension. \\
$Z_{\mathcal F},Z_i$ & Original--oracle observable pairs over $\mathcal Q$; $Z_i=Z_{\{u_i\}}$ for singleton deletion. \\
$\psi(Z_i),d$ & Shared evaluation vector and its feature dimension. \\
$r$ & Rank of the matrix of shared evaluation vectors. \\
$\rho=b/a$ & Allowed control-to-target score ratio in the selectivity analysis; distinct from FPR. \\
$\Phi,m,\kappa$ & Frozen expansion map, dictionary width, and SAE TopK sparsity. \\
$h_M(x)$ & SAE activations before frozen normalization. \\
\midrule
\multicolumn{2}{@{}l@{}}{\textit{Detection criteria and budgets}} \\
$s_i(M),\tau_i$ & Deletion score and its frozen threshold for target $i$. \\
$\alpha,\beta$ & FPR ceiling and TPR floor; defaults are $0.05$ and $0.90$. \\
$\widehat{\mathrm{TPR}}_i$ & Rate of flagging $u_i$ when the requested deletion is $u_i$. \\
$\widehat{\mathrm{FPR}}_{ji}$ & Rate of flagging $u_i$ when the requested deletion is $u_j$, $j\ne i$. \\
$\widehat{\mathrm{Cov}}_{\mathrm{obs}}$ & Fraction of targets satisfying both rate constraints in Eq.~(11). \\
$B,T_{\mathrm{opt}}$ & Responses per evaluation trial and optimizer updates per edit. \\
$L_{\mathrm{cal}},L_{\mathrm{test}}$ & Repeated calibration/test trials per model; distinct from the response budget $B$. \\
$\mathcal L$ & Loss function; distinct from trial counts denoted by $L$. \\
\bottomrule
\end{tabular}
\end{table*}

\subsubsection{Datasets and deletion targets}
\label{app:datasets_implementation}
\textbf{Compositional density task.}
The dataset contains every color--shape pair exactly once, with
$k\in\{4,8,16\}$ colors and shapes, giving $n\in\{16,64,256\}$ records.
Inputs are concatenated one-hot vectors. A singleton request removes one
pair from the training records; the complete set of pairs remains the
model's public output support. This construction controls which targets
share a color or a shape.

\textbf{Oxford-IIIT Pets.}
The image classification task uses 37 breed labels and 3,680 training
images~\cite{ceo_parkhi2012pets}. Each deletion target is one training image.

\textbf{FGVC Aircraft.}
The task uses 30 manufacturer labels~\cite{ceo_maji2013aircraft}.
The saved training pool contains 6,475 images after downsampling three
variants to three examples each. Each target is one training image.

\textbf{Language records.}
The two constructed QA tasks associate fictional people with access words,
or researchers with research stations. A singleton is one training QA;
four paraphrases of its question are reserved for evaluation. Deleting one
QA does not request removal of every fact associated with that person.
The model and observation comparisons for Figure~6 are described in
Section~\ref{app:figure6}.

\subsubsection{Implementation details}
\label{app:implementation}
\textbf{Synthetic models.}
A tanh network with layer widths $(2k,64,64,64,1)$ assigns one scalar
to each color--shape pair. A softmax over the complete
support defines $p_M$, including the pair omitted from training.
Original and Retraining start from matched initial weights and use
full-batch Adam for 800 updates at learning rate $0.003$.
Calibration oracle seeds are 1101--1102; test seeds are 2201--2203.

\textbf{Image models.}
The archived observability banks use ImageNet-initialized
ResNet-18~\cite{ceo_he2016resnet}. Both tasks use deterministic
$224\times224$ inputs, 24 epochs, batch size 64, SGD momentum $0.9$,
weight decay $5\times10^{-4}$, and backbone/head learning rates
$0.001/0.01$. All parameters are trainable, while batch-normalization
running moments stay fixed. A singleton retrain uses the matched
initialization and otherwise matched full-pool minibatches, excluding the
requested image. These oracles concern downstream training conditional on
the pretrained weights.

\textbf{Language adapters.}
LoRA retraining~\cite{ceo_hu2022lora} restarts an adapter from the same
pretrained base after excluding the requested QA. Access-word observations
use next-token likelihood; multi-token station answers use mean answer-token
NLL. Both are teacher-forced measurements. The four evaluation paraphrases
are not added to the training set.

\textbf{Unlearning defaults.}
Unless a method specifies otherwise, synthetic edits start from the saved
Original and use a fresh Adam optimizer for $T_{\mathrm{opt}}=80$
full-batch updates at learning rate $10^{-3}$. Section~\ref{app:baselines}
gives the method-specific objectives and exceptions. Figure~4 varies the
total update budget separately (Section~\ref{app:figure4}).

\textbf{Hyperparameter selection.}
The six recent-method settings were selected on a separate $k=8$, seed-17
development run by minimizing the mean of
\begin{equation}
\label{app:eq:development}
 J_{\mathrm{dev}}(M,u_i)
 =n p_M(u_i)+\frac{\mathrm{TV}_{\mathcal R_i}(M)}{0.01}
\end{equation}
over targets. Here $M$ is the model edited for target $i$,
$\mathcal R_i=\mathcal D\setminus\{u_i\}$, and retained-distribution TV
is defined in Eq.~\eqref{app:eq:retain_tv}. This selection uses
neither test seeds nor evaluator thresholds.

\subsubsection{Unlearning baselines}
\label{app:baselines}
We describe the 20 configurations in Table~1A in table order. Each entry
states the method's principle and its implementation for the color--shape
density task. Common optimization and hyperparameter selection follow
Section~\ref{app:implementation}; method-specific settings are given below.

For a requested record $u_i$, write $\ell_i=\log p_\theta(u_i)$ for the
current log-probability and $\ell_i^{\mathcal D}=\log p_{M_{\mathcal D}}(u_i)$
for its frozen Original reference. The common retain-set loss is
\begin{equation}
\label{app:eq:retain_nll}
 \mathcal L_{\mathrm{ret}}=-\frac{1}{n-1}\sum_{j\ne i}\log p_\theta(u_j).
\end{equation}
Objectives introduced for language or classification are adapted to
this categorical density.

\textbf{(1) Original.}
The no-edit reference minimizes empirical NLL on all $n$ records using
800 Adam updates at learning rate $0.003$. It returns the same parameters
for every request, providing a control in which no deletion is performed.

\textbf{(2) Retraining.}
Retain-set retraining provides the deletion reference~\cite{ceo_bourtoule2021unlearning}.
It restarts from Original's initial weights and minimizes
$\mathcal L_{\mathrm{ret}}$ for 800 Adam updates at learning rate $0.003$.
The target is excluded from training examples while remaining in the
public normalization support.

\textbf{(3) Finetune.}
Retain-set fine-tuning continues training on the records to be
preserved~\cite{ceo_kurmanji2023unbounded}. Our implementation minimizes
$\mathcal L_{\mathrm{ret}}$ in Eq.~\eqref{app:eq:retain_nll} from Original
with the common optimizer and budget. Retained records supply all positive
training examples.

\textbf{(4) GA.}
Gradient ascent suppresses the requested record by maximizing its
NLL~\cite{ceo_kurmanji2023unbounded,ceo_zhang2024npo}. We equivalently
minimize $\ell_i$ with the common optimizer and budget. The objective
contains only the forget term; density normalization uses the full
public support.

\textbf{(5) NPO.}
Negative Preference Optimization suppresses target likelihood relative
to a frozen reference~\cite{ceo_zhang2024npo}. For one categorical record,
we minimize
\begin{equation}
\label{app:eq:npo}
 \mathcal L_{\mathrm{NPO}}
 =\frac{2}{\beta_{\mathrm{NPO}}}
  \operatorname{softplus}\!\left(
   \beta_{\mathrm{NPO}}(\ell_i-\ell_i^{\mathcal D})\right).
\end{equation}
We use Original as the reference and set $\beta_{\mathrm{NPO}}=0.1$.
This configuration optimizes only the NPO objective.

\textbf{(6) NegGrad+.}
NegGrad+ combines target suppression with retain-set
training~\cite{ceo_kurmanji2023unbounded}. We minimize
\begin{equation}
\label{app:eq:neggrad}
 \mathcal L_{\mathrm{NegGrad+}}=0.95\mathcal L_{\mathrm{ret}}+0.05\ell_i.
\end{equation}
Both terms are evaluated in every update, with coefficients fixed across
targets and task sizes.

\textbf{(7) NPO+retain.}
This variant combines NPO in Eq.~\eqref{app:eq:npo} with retain-set
NLL~\cite{ceo_zhang2024npo}:
\begin{equation}
\label{app:eq:npo_retain}
 \mathcal L_{\mathrm{NPO+retain}}=0.05\mathcal L_{\mathrm{NPO}}+0.95\mathcal L_{\mathrm{ret}}.
\end{equation}
We use $\beta_{\mathrm{NPO}}=0.1$, a frozen Original reference, and the
same weights for every target. The requested record is excluded from
the retain term.

\textbf{(8) NPO+cKL.}
This density-specific NPO variant preserves Original's conditional
retain distribution~\cite{ceo_zhang2024npo}. We minimize
\begin{equation}
\label{app:eq:npo_ckl}
 \mathcal L_{\mathrm{NPO+cKL}}=0.05\mathcal L_{\mathrm{NPO}}+0.95\mathcal L_{\mathrm{cKL}}.
\end{equation}
The conditional retention loss is
\begin{equation}
\label{app:eq:conditional_kl}
 \mathcal L_{\mathrm{cKL}}=\mathrm{KL}\!\left(
 p_{M_{\mathcal D}}(\cdot\mid\mathcal R_i)
 \,\|\,p_\theta(\cdot\mid\mathcal R_i)\right).
\end{equation}
We set $\beta_{\mathrm{NPO}}=0.1$. The KL is computed over retained
color--shape pairs after renormalization, allowing the total target
mass to change independently of their relative probabilities.

\textbf{(9) SSD.}
Selective Synaptic Dampening reduces parameters disproportionately
important to the forget record~\cite{ceo_foster2024ssd}. We measure
importance by per-record squared NLL gradients at Original. Parameters
whose target importance exceeds ten times mean full-data importance are
multiplied by the full/target importance ratio, clipped at one. The
dampening coefficient is one; the edit requires no iterative optimizer.

\textbf{(10) Diag. Fisher.}
This local influence-based baseline approximates the effect of removing
one record using diagonal curvature~\cite{ceo_koh2017influence}. It applies
one elementwise update:
\begin{equation}
\label{app:eq:diagonal_fisher}
\begin{aligned}
 \theta'=\theta_{\mathcal D}
       +\frac{g_i-\bar g}{(n-1)(I+0.01)},
 \bar g=\frac1n\sum_jg_j,
 I=\frac1n\sum_jg_j^2.
\end{aligned}
\end{equation}
Here $g_j$ is the record-NLL gradient at Original; squares and division
are elementwise. The curvature approximation is the diagonal empirical
Fisher with damping $0.01$.

\textbf{(11) SimNPO+R.}
SimNPO removes the reference-model dependence from
NPO~\cite{ceo_fan2025simnpo}. Our single-event adaptation with retain NLL
minimizes
\begin{equation}
\label{app:eq:simnpo}
 \mathcal L_{\mathrm{SimNPO+R}}=2\operatorname{softplus}(\ell_i)+\mathcal L_{\mathrm{ret}}.
\end{equation}
This uses $\beta_{\mathrm{SimNPO}}=1$, offset zero, and equal forget/retain
weights. Each generated combination has length one, so the sequence-length
normalization is one.

\textbf{(12) WGA+R.}
Weighted Gradient Ascent scales the forget gradient by the current
target probability~\cite{ceo_wang2025wga}. With retain NLL, we minimize
\begin{equation}
\label{app:eq:wga}
 \mathcal L_{\mathrm{WGA+R}}
 =\operatorname{stopgrad}\!\left(p_\theta(u_i)^{\beta_{\mathrm{WGA}}}\right)
   \ell_i+\mathcal L_{\mathrm{ret}}.
\end{equation}
We set $\beta_{\mathrm{WGA}}=1$ and use equal forget/retain weights.
The probability weight is detached when computing the gradient.

\textbf{(13) UNDIAL.}
UNDIAL distills a frozen teacher after reducing the target
logit~\cite{ceo_dong2025undial}. On the public support, our teacher and
student objective are
\begin{equation}
\label{app:eq:undial}
\begin{aligned}
 q_j\propto\exp(\ell_j^{\mathcal D}-3\delta_{ij}),
 \mathcal L_{\mathrm{UNDIAL}}=-\sum_jq_j\log p_\theta(u_j).
\end{aligned}
\end{equation}
Here $\delta_{ij}$ is one only for the requested target. The teacher
starts from Original, reduces the target logit by $3$, and renormalizes
over all records. Distillation supplies the entire objective.

\textbf{(14) RMU.}
Representation Misdirection for Unlearning redirects forget activations
while preserving retain activations~\cite{ceo_li2024wmdp}. We apply it to
the final 64-dimensional tanh activation, minimizing target MSE to a
fixed random unit vector plus ten times retain activation MSE to Original.
Only the final hidden affine layer is updated; the energy head is frozen.
Control-vector seeds are $20261114+1009i+97k$.

\textbf{(15) NegMerge.}
NegMerge negates the sign-consistent part of several target-fine-tuning
vectors~\cite{ceo_kim2025negmerge}. Three copies of Original minimize
target NLL for 80 updates each at learning rates $0.0005$, $0.001$, and
$0.002$. We average parameter changes where all signs agree and subtract
the result from Original with scale one. Its total cost is 240 updates.

\textbf{(16) Saliency-NegGrad+.}
This component control combines SalUn's gradient-saliency
mask~\cite{ceo_fan2024salun} with the NegGrad+
objective~\cite{ceo_kurmanji2023unbounded}. We select the largest $10\%$
of absolute target-NLL gradients at Original, freeze this parameter mask,
and optimize Eq.~\eqref{app:eq:neggrad} within it. Thus the control
uses SalUn's parameter selection with our specified forget/retain loss.

\textbf{(17) Amnesiac rollback.}
Amnesiac unlearning subtracts training updates involving the forgotten
record~\cite{ceo_graves2021amnesiac}. Every original update is full-batch,
so all 800 updates involve the target. We reconstruct their cumulative
change from saved initial and final parameters; subtracting it restores
initialization exactly, with no further optimization.

\textbf{(18) Amnesiac+repair.}
This local control repairs the model after Amnesiac
rollback~\cite{ceo_graves2021amnesiac}. It minimizes
$\mathcal L_{\mathrm{ret}}$ for 80 updates at learning rate $0.001$ using
a fresh Adam optimizer. Here rollback restores initialization, so repair
is retain training from initialization with this reduced budget.

\textbf{(19) BlindSpot-event.}
Bad-teacher distillation uses different teachers for forget and retain
examples~\cite{ceo_chundawat2023badteacher}. Our event adaptation uses
Bernoulli distributions $[p_\theta(u),1-p_\theta(u)]$: the saved untrained
network teaches the requested target, and Original teaches every retained
record. We minimize teacher-to-student KL, averaged over records and
outcomes, at temperature one.

\textbf{(20) FullTeacher.}
This control for BlindSpot-event~\cite{ceo_chundawat2023badteacher} uses
Original as teacher for every record, including the requested target.
The event interface, KL direction, temperature, optimizer, and budget
match BlindSpot-event. Their comparison isolates the effect of replacing
the target's teacher.

\subsubsection{Evaluation protocol and metrics}
\label{app:deletion_trials}
\textbf{Positive trials and controls.}
For queried target $u_i$, a positive trial requests deletion of $u_i$;
a negative control requests deletion of $u_j$, $j\ne i$, retaining $u_i$.
These labels indicate actual training-set exclusion for retraining oracles
and the requested edit for approximate methods. A deletion is flagged when
$s_i(M)>\tau_i$, with ties counted as negative. Thresholds use only
calibration oracles and remain frozen for all tested unlearning methods.

\textbf{Target coverage.}
Empirical coverage follows Eq.~(11): it is the fraction of the $n$ targets
that satisfy both
\begin{equation}
\label{app:eq:coverage_gate}
 \widehat{\mathrm{TPR}}_i\ge\beta,
 \qquad
 \max_{j\ne i}\widehat{\mathrm{FPR}}_{ji}\le\alpha.
\end{equation}
The default requirements are $\beta=0.90$ and $\alpha=0.05$.
Mean TPR averages targets. Max.\ FPR takes the maximum over ordered
target--control pairs within a run. Table~1 reports the mean and sample
standard deviation of these run-level statistics over three training seeds.

\textbf{Repetitions and uncertainty.}
Repeated output draws estimate conditional rates for a fixed model;
training seeds quantify variation between learned models. For image case
banks with a single target-excluded model, we report detection and empirical
false alarms rather than treating one positive as repeated oracle evidence.
Confidence-supported coverage, reported separately in Appendix~C,
requires simultaneous lower TPR and upper FPR bounds to meet
Eq.~\eqref{app:eq:coverage_gate}.

\textbf{Distribution quality.}
For the compositional density task, request coverage is accompanied by
normalized target mass $n p_M(u_i)$ and retained-distribution error.
For $\mathcal R_i=\mathcal D\setminus\{u_i\}$, the latter is
\begin{equation}
\label{app:eq:retain_tv}
 \mathrm{TV}_{\mathcal R_i}(M)=\frac12\sum_{u\in\mathcal R_i}
 \left|p_M(u\mid\mathcal R_i)-\frac1{n-1}\right|.
\end{equation}
These quality measures use the full modeled distribution, independently of
sampled detector decisions.

\subsubsection{Frozen expansion and scoring}
\label{app:sae_expansion}
\textbf{SAE fitting.}
An SAE encodes each response according to Eq.~(12), with dictionary width
$m$ and TopK sparsity $\kappa$. Its decoder and reconstruction objective are
\begin{equation}
\label{app:eq:sae_objective}
 \begin{aligned}
 \widehat z_{M_{\mathcal D}}(x)
   =A_d h_{M_{\mathcal D}}(x)+b_d,
 \mathcal L_{\mathrm{SAE}}
 =\mathbb E_x\bigl\|z_{M_{\mathcal D}}(x)
                     -\widehat z_{M_{\mathcal D}}(x)\bigr\|_2^2.
 \end{aligned}
\end{equation}
The dictionary and normalization are fitted on Original activations and
then frozen across all edits.

\textbf{Paired scoring.}
Equation~(13) compares SAE-only activations by RMS distance.
The corresponding raw-embedding score is
\begin{equation}
\label{app:eq:raw_score}
 s_i^{\mathrm{raw}}(M)=
 \frac{\|z_M(u_i)-z_{M_{\mathcal D}}(u_i)\|_2}{\sqrt p}.
\end{equation}
Concatenating raw and SAE features defines a separate observable.
Each representation receives its own threshold on the same calibration
controls and is tested on the same model pairs and queries.

\textbf{Synthetic mechanism control.}
This learned encoder is distinct from the public-support random-feature
decoder used for the synthetic mechanism experiment in
Section~\ref{app:table3}. Deterministic expansion changes which response
relationships a fixed scoring rule can access; it does not add observations.

\subsection{Settings for individual tables and figures}
\label{app:display_settings}
Figures~1 and 3 illustrate the evaluation framework; the experimental
comparisons are specified below. Figure~3's four-pair example is separate
from the $16/64/256$-target results in Table~1.
\subsubsection{Table 1A: comparing unlearning methods}
\label{app:table1_methods}
\textbf{Comparison.}
Table~1A compares unlearning methods with a fixed evaluator. All 20
configurations use Joint Frequency on every singleton request, at
$n\in\{16,64,256\}$. Each evaluation trial draws $B=4(n-1)$ outputs,
giving 60, 252, and 1,020 samples, respectively. The model family, target
set, evaluator, and observation budget are shared across methods.
Original is the no-edit control; Retraining is the deletion oracle.
Section~\ref{app:baselines} gives the update procedures, including
exceptions to the default 80-update budget.

\textbf{Shared calibration.}
For each evaluator, task size, and budget, we use
$L_{\mathrm{cal}}=256$ output trials per deletion identity per calibration
seed, and $L_{\mathrm{test}}=1{,}024$ per test model. Calibration negatives
are pooled across targets in three groups: same color, same shape, or
neither. For a group of $q_g$ scores, use the one-indexed order statistic
\begin{equation}
\label{app:eq:calibration_order}
 r_g=\min\bigl\{q_g,\lceil0.95(q_g+1)\rceil\bigr\}.
\end{equation}
The largest group cutoff becomes one shared threshold. Scores are rounded
to 12 decimal places before strict comparisons. Coverage is then computed
by checking every target--control pair on the test trials, rather than
assuming that the pooled calibration controls satisfy the per-target gate.

\subsubsection{Table 1B: comparing evaluators on fixed oracles}
\label{app:table1_evaluators}
\textbf{Comparison.}
Table~1B isolates the evaluator: all six configurations receive the same
sampled outputs from each Retraining model and follow the calibration
protocol in Section~\ref{app:table1_methods}. For target $(a,b)$, let
$N_{ab}$ be its joint count. The marginal counts are
$N^c_a=\sum_bN_{ab}$ and $N^s_b=\sum_aN_{ab}$.
These are evaluation statistics, not the density MLP's hidden embedding.

\textbf{Marginal Linear and Redundant Linear.}
The raw score is
\begin{equation}
\label{app:eq:marginal_linear}
 s_{ab}^{\mathrm{linear}}=\frac2k-\frac{N^c_a+N^s_b}{B}.
\end{equation}
Redundant Linear isometrically embeds the $2k$ marginal vector into $4k$
coordinates and recovers Eq.~\eqref{app:eq:marginal_linear}, isolating
coordinate redundancy.

\textbf{Marginal Quadratic.}
We standardize the two counts using their uniform binomial means and
variances to obtain $x,y$, then fit logistic regression on five features:
\begin{equation}
\label{app:eq:quadratic_features}
 f(x,y)=\bigl[1,\ x+y,\ xy,\ x^2+y^2,\ (x-y)^2\bigr],
\end{equation}
with $\ell_2$ penalty $10^{-3}$.
Fitting uses independent output streams from the calibration-model seeds;
neither the feature family nor its penalty is chosen on test outcomes.

\textbf{Marginal Likelihood.}
We normalize the compatibility weights
\begin{equation}
\label{app:eq:marginal_likelihood}
 H_{ab}=\binom{B-N^c_a}{N^s_b}\big/\binom{B}{N^s_b}
\end{equation}
over all pairs. The score exploits the public uniform construction without
reading the sampled pair counts.

\textbf{Joint Frequency and Zero Count.}
Joint Frequency uses a calibrated threshold on
\begin{equation}
\label{app:eq:joint_score}
 s_{ab}^{\mathrm{joint}}=\frac1n-\frac{N_{ab}}B.
\end{equation}
Zero Count flags deletion exactly when $N_{ab}=0$; it keeps this fixed rule.
Both read the same joint counts, so their comparison isolates the decision
rule rather than the observation interface.

\subsubsection{Figure 2: image observability and target diagnosis}
\label{app:figure2}
This experiment varies which directions of the fixed model responses are
observed. The case analysis below specifies how a previously missed
oracle deletion is diagnosed as the observation width increases.

\textbf{Model responses and nested widths.}
The 100-target image banks retain every sampled target and use the
ResNet-18 training settings in Section~\ref{app:implementation}.
The 512D embeddings are orthogonally aligned to Original by Procrustes
using 1,024 independent inputs. A shared nested basis starts with the
centered Original head span (36 Pets or 29 Aircraft directions), then
adds residual principal components fitted on those inputs. A single image
forward pass provides every nested width. Native logits use each model's
own head and are evaluated as a separate interface.

\textbf{Case scoring and calibration.}
For each Pets target, 24 target-retaining models estimate coordinate means
and variances; 25 different retaining models calibrate the maximum-score
threshold; 50 held-out retaining models estimate false alarms. The score
is the largest squared standardized coordinate deviation. Its variance
floor is $1\%$ of the mean member variance over all 512 coordinates and
stays fixed across widths. The target-excluded model supplies one positive.

\textbf{Detection and false alarms in panel (b).}
Case A (Great Pyrenees) was detected at all ten widths, with no false
alarms among the 50 held-out controls. Case B (English cocker spaniel)
was first detected at 256D, with two false alarms ($4\%$), and had the
same outcome at 320D. It remained detectable at 384D, 448D, and 512D,
but the respective false-alarm rates of $6\%$, $8\%$, and $8\%$ exceeded
the $5\%$ limit. Case C (American bulldog) was missed at every width
and exceeded the limit only at 384D ($6\%$). Each detection records the
outcome for one target-excluded model, so these cases do not estimate
TPR across independent retrainings.

Panel (c) shows the deviation along each direction for case B. Its
t-SNE inset visualizes the Original and retrained representations.

\subsubsection{Figure 4: edit budgets and non-target confusion}
\label{app:figure4}
\textbf{Budget sweep, panels (a--c).}
Panels (a--c) use $n=16,64,256$, respectively, at the corresponding
Table~1 budgets $B=4(n-1)$. The horizontal axis is total optimizer updates
$T_{\mathrm{opt}}\in\{30,60,90,120,150\}$, not epochs or model queries.
All checkpoints use the frozen historical Joint Frequency thresholds and
1,024 trials per model. Retraining is the same fixed 800-update reference
at every displayed budget.

\textbf{Method settings.}
Hyperparameters are inherited from the separate 800-update development
configurations and remain fixed throughout the sweep; UNDIAL uses penalty
10 and NegMerge scale $0.5$. NegMerge assigns $T_{\mathrm{opt}}/3$ updates to each of its
three candidates, so $T_{\mathrm{opt}}$ counts their total work. The legends ``NPO'' and
``Amnesia'' denote NPO+retain and Amnesiac+repair, respectively.

\textbf{Shared-attribute errors, panel (d).}
Panel (d) uses $T_{\mathrm{opt}}=60$ and groups false alarms by whether the non-target
deletion shares the queried color or shape. Of the $n-1$ alternatives,
$2(k-1)$ share one attribute: $30/255=11.76\%$ at $k=16$.
The plotted error share is the fraction of false-positive decisions in
this group, not that group's FPR.

\subsubsection{Figure 5 and Table 2: score selectivity and partial deletion}
\label{app:linear_controls}
These comparisons test the linear-score conditions in Section~4.
The response ratio $\rho$ measures selectivity on a normalized score
scale; it is distinct from the empirical FPR used for target coverage.

\textbf{Figure 5: score scale and rank.}
Each target's response is normalized to one on its own oracle deletion;
other scores are relative responses on this scale, not detection rates.
Centered separate counts have rank $r=2k-2$, while centered joint counts
have rank $r=n-1$. Rearranging Eq.~(6) gives the necessary floor
\begin{equation}
\label{app:eq:rank_floor}
 \rho\ge\sqrt{\frac{n-r}{r(n-1)}}.
\end{equation}
For $n=256,r=30$, it is $17.19\%$. The relation-group analysis separates
controls sharing the target's color, shape, or neither; the diagonal is
the target's own deletion. The rank floor is a necessary condition for
linear selectivity and is not a coverage or FPR bound.

\textbf{Table 2: controlled subset comparison.}
Table~2 counts the requests within the $25\%$ score limit for the
\emph{Uniform} oracle. Its \emph{Retrained} column reports the range of
worst-control scores for learned retraining references. For the uniform
oracle, $n=16$. An index subset $T$ specifies
the deleted records $\mathcal F_T=\{u_j:j\in T\}$ and response
$v_T=p_{\mathcal D}-p_{\mathcal D\setminus\mathcal F_T}$.
For each requested index set $S$, solve
\begin{equation}
\label{app:eq:selectivity_lp}
 \begin{aligned}
 \rho_S^*&=\min_{w\in\mathbb R^d,\ t\ge0}t,\\
 \text{s.t.}\quad w^\top v_S=1, & \quad -t\le w^\top v_T\le t\quad(T\in\mathcal C_S).
 \end{aligned}
\end{equation}
We enumerate all $\binom{16}{2}=120$ pairs or
$\binom{16}{3}=560$ triples. Pair requests have 91 disjoint and 28 partial
controls; the matched-count comparison replaces 28 disjoint controls with
the 28 partial controls, retaining 91 total. Triple requests have 286,
520, or 559 controls when admitting zero, at most one, or at most two
shared members.

\textbf{Table 2: reported quantities.}
The uniform optima are $1/7,3/7,3/7$ for pairs and
$3/13,3/13,23/39$ for triples. Under the fixed $\rho_S^*\le0.25$ criterion,
the uniform oracle passes $120/120$, $0/120$, and $0/120$ pair requests,
and $560/560$, $560/560$, and $0/560$ triple requests, in table order.

\subsubsection{Table 3: expansion before versus after aggregation}
\label{app:table3}
\textbf{Controlled comparison.}
For each specified feature map in Table~3, Pre and Post use the same
map and the same linear readout. Only the order of mapping and averaging
changes. The comparison fixes $n=256$, $B=1{,}020$, all model pairs,
and every sampled output.
Write a response as $z_t=[c_t;s_t]$, where $c_t,s_t$ are its color and
shape one-hot vectors. The two expansion maps are
\begin{equation}
\label{app:eq:expansion_maps}
 \begin{aligned}
 \phi_{\mathrm{ReLU}}(z_t)
   &=\bigl[z_t;\mathrm{ReLU}(Wz_t+b)\bigr],\\
 \phi_{\mathrm{int}}(z_t)
   &=\bigl[z_t;\operatorname{vec}(c_ts_t^\top)\bigr].
 \end{aligned}
\end{equation}
Both expanded representations have width 288 including 32 raw coordinates.
For either map $\phi$, the aggregation orders are
\begin{equation}
\label{app:eq:aggregation_order}
 \begin{aligned}
 \bar z&=\frac1B\sum_{t=1}^{B}z_t,\\
 \mathrm{Pre}&=\frac1B\sum_{t=1}^{B}\phi(z_t),
 &\mathrm{Post}&=\phi(\bar z).
 \end{aligned}
\end{equation}
\textbf{Decoder and repetitions.}
The map and decoder are shared: the decoder is the pseudoinverse of the
centered public-support feature table, with relative cutoff $10^{-9}$,
and the two representations are centered at their respective uniform
references. The resulting Pre/Post contrast measures what aggregation
preserves for this map and public-support decoder. The frozen SAE in
Section~\ref{app:sae_expansion} uses a learned dictionary and a different score.
ReLU uses map seeds 3401 and 3402, crossed with the three model seeds;
its table entries average all six runs. Other entries average three model seeds.

\textbf{Expansion-study calibration.}
This study uses $L_{\mathrm{cal}}=512$ per deletion per calibration seed
and $L_{\mathrm{test}}=1{,}024$ per test model. For each queried target,
take the largest of the three relation-group 95th percentiles, using the
``higher'' quantile convention. Scores are rounded to a $2^{-32}$ grid
before strict comparisons. Unlike Table~1, thresholds are target-specific
because random maps break symmetry.

\textbf{Reported quantities and controls.}
Table~3's FPR column averages non-target deletions; coverage checks the
worst control for every target. Raw (32D), redundant Linear (288D), and
Joint counts (256D) use the same output draws as the expanded
representations.

\subsubsection{Figure 6: SAE fitting data and dictionary width}
\label{app:figure6}
Figure~6 uses four model--data settings: Qwen2.5-3B and Qwen3-1.7B for
language-fact deletion, ResNet-18 on FGVC Aircraft, and DINOv2 on
Oxford-IIIT Pets. In each panel, target coverage is plotted against the
SAE dictionary-to-embedding width ratio $m/p\in\{16,64,256\}$, with
separate curves for dictionaries fitted on $12.5\%$, $25\%$, $50\%$, or
$100\%$ of the Original-model data. Curves include raw-embedding coverage
as their reference. The main text applies a frozen SAE to the same language
and vision embeddings and includes a Random TopK control matched to the SAE
in width and sparsity. Equation~(13) defines the paired score on SAE
activations.

\subsubsection{Figure 7: query budget and coverage criteria}
\label{app:figure7}
\textbf{Query-budget sweep, panel (a).}
The frozen-model study reuses the $n=256$ retraining families with
$B\in\{1{,}020,1{,}530,8{,}192,65{,}536\}$, using the calibration
procedure in Section~\ref{app:table3}. Each budget generates new output
draws from the same saved models; it does not add training seeds.
Within a budget, Joint Frequency and Zero Count read identical pair counts.

\textbf{Coverage-criterion sweep, panel (b).}
We recompute Eq.~(11) from the saved per-target TPR
and worst FPR while varying $\alpha$ and $\beta$. Turning a criterion
``off'' removes only that gate. Detector scores and calibrated decision
thresholds remain fixed; this is a coverage-criterion sweep, not detector
recalibration. The selected query budget must be held fixed across the
compared heatmaps.

\section{Proofs for the Capacity Limits of Section~\ref{sec:diag}}
\label{app:capacity-analysis}

\subsection{Capacity limits for singleton coverage}
\label{app:singleton-capacity}

\begin{proof}[Proof of Theorem~1]
Fix the oracle observations and the shared representation used by the
evaluator. Write $v_i=\psi(Z_i)$, and let column $j$ of $W$ contain
the coefficients $w_j$ of the score for target $u_j$.
Thus $C_{ij}=v_i^\top w_j$, $C=V^\top W$, and
$\operatorname{rank}(C)\le r=\operatorname{rank}(V)\le d$.
Suppose all $n\ge 2$ targets meet the selectivity requirement at
levels $a>b\ge 0$, and set $\rho=b/a$.

Since $C_{jj}\ge a>0$, the diagonal matrix
$D=\operatorname{diag}(C_{11},\ldots,C_{nn})$ is invertible.
Dividing each score column by its own diagonal entry gives $B=CD^{-1}$.
Then $B_{jj}=1$, $\lvert B_{ij}\rvert\le \rho$ for $i\ne j$,
and $\operatorname{rank}(B)=\operatorname{rank}(C)$.
For a real square matrix with positive singular values $\sigma_\ell$,
the singular value decomposition and Cauchy--Schwarz give
\begin{equation}
  \lvert\operatorname{tr} B\rvert
  \le \sum_\ell \sigma_\ell
  \le \sqrt{\operatorname{rank}(B)}\,\lVert B\rVert_{\mathrm F}.
\end{equation}
Using $\operatorname{tr}B=n$ and summing the squared entry bounds yields
\begin{equation}
  n^2
  \le \operatorname{rank}(B)\,\lVert B\rVert_{\mathrm F}^2
  \le r\bigl(n+n(n-1)\rho^2\bigr).
\end{equation}
Rearranging gives the first bound of Theorem~1,
\begin{equation}
  d\ge r\ge \frac{n}{1+(n-1)\rho^2}.
\end{equation}
When $n^{-1/2}\le\rho<1/2$, Alon's perturbed-identity rank
theorem~\cite[Theorem~2.1]{alon2009perturbed} applies directly to $B$.
For a universal constant $c>0$ it gives
\begin{equation}
  d\ge r\ge \operatorname{rank}(B)
       \ge c\,\frac{\log n}{\rho^2\log(1/\rho)}.
\end{equation}
At $\rho=0$, $B=I_n$, so $r=n$.
\end{proof}

\noindent\textbf{Implication for Finding~3.}
For fixed $0<\rho<1/2$, the logarithmic bound applies once
$n\ge\rho^{-2}$. A representation below the required rank cannot
support all $n$ selective scores, even when every deletion uses its
retraining oracle. The bounds give a necessary capacity condition.

\subsection{Capacity for arbitrary subset coverage}
\label{app:subset-capacity}

\begin{proof}[Proof of Corollary~2]
Fix $1\le K\le n/2$ and let $\mathcal S_K$ be the family of all
$K$-element subsets of $[n]$, with $M=\binom nK$.
Index the rows of a score matrix by the deletion $T\in\mathcal S_K$
and its columns by the request $S\in\mathcal S_K$. With
$v_T=\psi(Z_T)$, its entries are
\begin{equation}
  C^{(K)}_{T,S}=v_T^\top w_S .
\end{equation}
This matrix factors through the shared $d$-dimensional representation,
so its rank is at most $d$. Selectivity for every request gives a
diagonal entry of at least $a$ and off-diagonal magnitudes of at most
$b$. These comparisons include every $T\ne S$ of size $K$,
including deletions that overlap $S$.

Applying the singleton result to these $M$ requests yields
$d\ge c\log M/[\rho^2\log(1/\rho)]$ when
$M^{-1/2}\le\rho<1/2$. For $0\le j<K$, the inequality
$(n-j)/(K-j)\ge n/K$ gives
\begin{equation}
  M=\prod_{j=0}^{K-1}\frac{n-j}{K-j}
    \ge (n/K)^K .
\end{equation}
Substituting $\log M\ge K\log(n/K)$ gives the stated bound
\begin{equation}
  d\ge c\,\frac{K\log(n/K)}{\rho^2\log(1/\rho)} .
  \qedhere
\end{equation}
\end{proof}

\subsection{When controls preclude subset selectivity}
\label{app:subset-controls}

\begin{proof}[Proof of Proposition~3]
Fix $S\in\mathcal S_K$, a control family
$\mathcal C_S\subseteq\mathcal S_K\setminus\{S\}$, and levels
$a>b\ge 0$. All sums in this proof run over $T\in\mathcal C_S$.
A feasible score has coefficients $w\in\mathbb R^d$ satisfying
\begin{equation}
  w^\top v_S\ge a,\qquad
  -b\le w^\top v_T\le b\quad (T\in\mathcal C_S).
\end{equation}

First suppose $v_S=\sum_T\lambda_T v_T$ for real coefficients
with $b\sum_T|\lambda_T|<a$. Every $w$ satisfying the control
constraints then obeys
\begin{equation}
  |w^\top v_S|
  \le \sum_T|\lambda_T|\,|w^\top v_T|
  \le b\sum_T|\lambda_T|<a .
\end{equation}
This rules out a feasible target score.

Conversely, suppose the constraints are infeasible. Write the target
constraint as $-v_S^\top w\le-a$, and the two inequalities for
each control as $v_T^\top w\le b$ and $-v_T^\top w\le b$.
Farkas' lemma gives nonnegative multipliers
$\eta,\mu_T^+,\mu_T^-$ such that
\begin{equation}
\begin{aligned}
  -\eta v_S+\sum_T(\mu_T^+-\mu_T^-)v_T &=0,\\
  -a\eta+b\sum_T(\mu_T^++\mu_T^-) &<0 .
\end{aligned}
\end{equation}
The second inequality forces $\eta>0$, since $b\ge 0$.
Set $\lambda_T=(\mu_T^+-\mu_T^-)/\eta$. The first equality
reconstructs $v_S=\sum_T\lambda_T v_T$, and
\begin{equation}
  b\sum_T|\lambda_T|
  \le \frac b\eta\sum_T(\mu_T^++\mu_T^-)<a .
\end{equation}
This proves the converse, including $b=0$.
\end{proof}

\noindent\textbf{Overlapping controls in Finding~4.}
If a request is selective against its disjoint controls, those controls
alone admit no reconstruction satisfying the strict bound above.
By Proposition~3, adding overlapping controls destroys selectivity
exactly when the enlarged family admits such a reconstruction. At least
one newly included control must then have a nonzero coefficient.
In the example $S=\{1,2\}$, if $v_S$ is the midpoint of
$v_{\{1,3\}}$ and $v_{\{2,3\}}$, the two coefficients are
$1/2$. Their absolute values sum to one, so $b<a$ certifies
infeasibility.

\smallskip
\noindent\textbf{Appending observables in Finding~4.}
Fix the enlarged control family and suppose it makes $S$ infeasible.
Write $\Lambda_S$ for the set of all signed coefficient vectors that
reconstruct $v_S$ from these controls and satisfy
$b\sum_T|\lambda_T|<a$. After appending coordinates $z_T$, apply
Proposition~3 to the joint vectors $(v_T,z_T)$.
A reconstruction of the joint target must use the same coefficients in
both coordinate blocks. Thus an obstruction survives exactly when
there is a $\lambda\in\Lambda_S$ with
\begin{equation}
   z_S=\sum_{T\in\mathcal C_S}\lambda_T z_T .
\end{equation}
Selectivity is restored exactly when every $\lambda\in\Lambda_S$
fails this equality on at least one appended coordinate.
\section{Supplementary Empirical Analyses}
\label{app:supplementary_results}
We examine how aggregation order affects a fixed readout, how validation
sample size affects the evidence for coverage, and how target mass and
distributional error help interpret coverage under a supplementary protocol.
\subsection{Aggregation order with a fixed public-support decoder}
\label{app:width_structure}
\begin{figure*}[t]
 \centering
 \includegraphics[width=\textwidth]{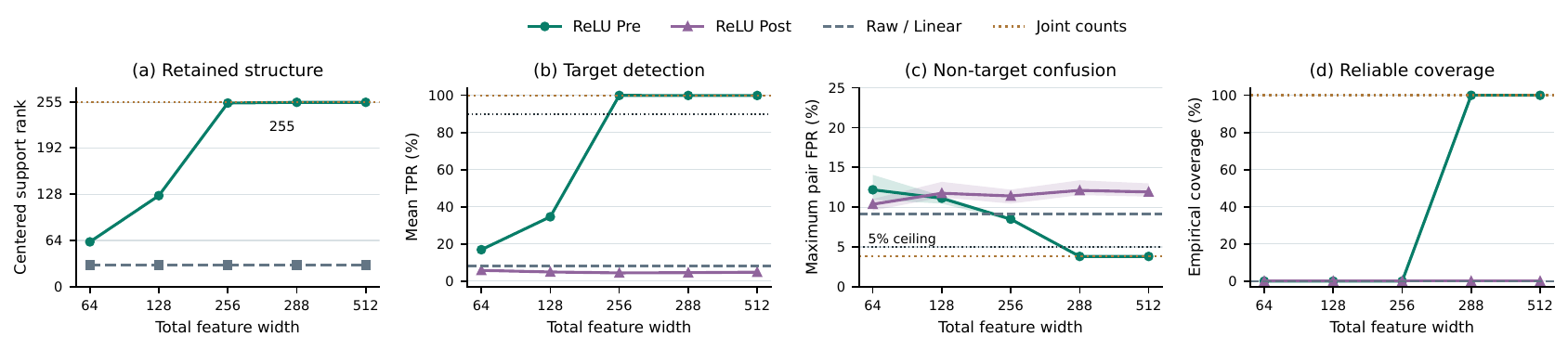}
 \caption{\textbf{Aggregation order and empirical coverage on the color--shape task.}
 All evaluators receive identical oracle output draws at $n=256$ and
 $B=1{,}020$. At each width and map seed, ReLU Pre and Post share the map
 and its public-support pseudoinverse. Each has its own calibrated thresholds.
 ReLU bands span the six runs from two map seeds and three model seeds.
 Horizontal Raw (32D) and Joint (256D) references show three-seed means.
 Linear expansion reproduces Raw. Panel (a) reports the rank of the centered
 public-support feature table. Panel (c) averages the maximum pair FPR from
 each run. Panel (d) counts targets whose empirical TPR is at least $90\%$
 and whose worst empirical FPR is at most $5\%$.}
 \label{fig:app-width}
\end{figure*}

Table~3 compares applying a fixed map before averaging (Pre) or after
averaging (Post) at width 288. Figure~\ref{fig:app-width} repeats this
comparison at five widths. Every map retains the 32 raw coordinates. The
centered ReLU feature table has rank 62 at width 64 and reaches rank 255
at widths 288 and 512. The redundant linear maps have rank 30 throughout.
At rank 255, the public-support pseudoinverse recovers the centered joint
counts from the mean of the mapped responses. Pre therefore gives the
same decisions as Joint. Post has access only to the mean color and shape
counts. At both widths Pre empirically covers all 256 targets in each run,
and Post and the redundant linear maps cover none. This mechanism concerns
the fixed maps and readouts on the finite color--shape support.

False alarms explain the transition from width 256 to 288. At width 256,
every target passes the empirical TPR gate (mean $99.93\%$), but each has
at least one control above the FPR ceiling. At width 288, mean TPR is
$99.89\%$ and every control meets the ceiling. The mean of the run-wise
maximum pair FPRs falls from $8.50\%$ to $3.81\%$. All 256 targets thus
pass the empirical point-estimate gates at width 288. Using the same
decisions and 1,024 validation trials per model, none passes the
simultaneous confidence-bound gates. These are two evidence criteria for
the same conditional TPRs and FPRs: one uses observed rates, and the other
requires simultaneous bounds to satisfy the thresholds. The difference
does not reflect a changed detector or conditional rate.

\subsection{Validating the coverage measurement}
\label{app:measurement}
\begin{figure*}[t]
 \centering
 \includegraphics[width=.94\textwidth]{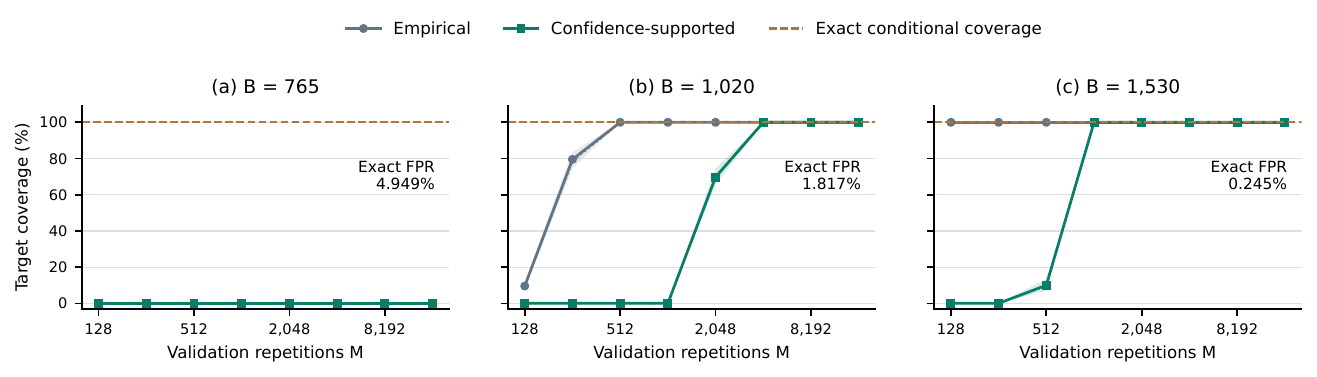}
 \caption{\textbf{Validation repetitions and evidence for target coverage.}
 The uniform leave-one-out oracle and Zero Count rule have exact conditional
 coverage $100\%$ at each displayed budget. Within a panel, only the number
 $M$ of validation repetitions changes. Solid curves average twelve independently
 sampled validation banks, with shading for their observed range. Dotted curves
 show the exact expectation of each coverage estimator. The horizontal dashed
 line is the known conditional coverage. These repetitions quantify sampling
 uncertainty for the fixed frequency models.}
 \label{fig:app-measurement}
\end{figure*}

We validate the coverage estimator against an oracle with known probabilities.
For $n=256$ uniformly weighted combinations, Zero Count always detects the
deleted target. When another target is deleted, its false-positive probability
is $p_B=(1-1/255)^B$. The three budgets in
Figure~\ref{fig:app-measurement} therefore have exact FPRs of $4.949\%$,
$1.817\%$, and $0.245\%$, all below the $5\%$ ceiling. Within each panel,
the budget, rule, and conditional rates are fixed. Empirical coverage
applies the $90\%/5\%$ gates to observed rates. Confidence-supported coverage
applies those same gates to a simultaneous lower bound on target TPR and an
upper bound on each control FPR, using the same decisions. Thus the two
coverage columns use an empirical point-estimate criterion and a simultaneous
confidence-bound criterion, respectively; they do not represent different
detectors or underlying conditional rates.

The expected empirical coverage is
\[
 \left[\Pr\{\mathrm{Binomial}(M,p_B)\leq\lfloor0.05M\rfloor\}\right]^{255}.
\]
The output streams for distinct deletions are independent, while targets
scored on a common stream can be dependent. The expression gives the
expected covered fraction. The simulation uses the joint empty-category occupancy law and was
checked against direct output sampling. Simultaneous Clopper--Pearson bounds
use one-sided tail probability $\delta=0.05/[2(n^2+4n)]$, the conservative
allocation retained from the archived protocol.

At $B=1{,}020$ and $M=512$, all 256 targets pass the empirical gates in
each validation bank, yet none passes the simultaneous-bound criterion.
At $M=2{,}048$ the bounds support $69.76\%$ of targets on average, and
at $M=4{,}096$ they support all targets in every bank. Only the amount of
validation evidence changes across these comparisons. At $B=765$ the
true FPR is $4.949\%$, just below the $5\%$ ceiling. At every tested $M$,
each target has at least one of its 255 controls exceed the empirical ceiling
in every bank, even though all targets satisfy the true conditional criterion.

\subsection{Interpreting request coverage under the Zero Count protocol}
\label{app:quality}
\begin{table*}[t]
\centering
\small
\setlength{\tabcolsep}{4.2pt}
\begin{tabular}{lrrrrrrr}
\toprule
Method & Target mass $\downarrow$ & Retain TV $\downarrow$ & Retrain TV $\downarrow$ & TPR & Max. FPR & Emp. cov. & Supp. cov. \\
\midrule
Original & 100.000 & 0.003 & 0.395 & 0.25 & 1.27 & 0.00 & 0.00 \\
Retraining & 0.026 & 0.027 & 0.000 & 99.84 & 1.37 & 100.00 & 100.00 \\
FT-80 & 58.120 & 0.495 & 0.551 & 4.88 & 1.37 & 0.00 & 0.00 \\
FT-800 & 0.052 & 0.042 & 0.053 & 99.69 & 1.17 & 100.00 & 100.00 \\
GA & $<0.001$ & 11.745 & 11.744 & 100.00 & 100.00 & 0.00 & 0.00 \\
NegGrad+ & $<0.001$ & 2.113 & 2.113 & 100.00 & 10.55 & 99.61 & 82.03 \\
SSD & 100.023 & 0.007 & 0.396 & 0.26 & 1.07 & 0.00 & 0.00 \\
NPO & $<0.001$ & 11.746 & 11.745 & 100.00 & 100.00 & 0.00 & 0.00 \\
NPO+R & $<0.001$ & 2.500 & 2.500 & 100.00 & 12.11 & 97.79 & 50.78 \\
NPO+cKL & $<0.001$ & 2.426 & 2.425 & 100.00 & 11.23 & 98.83 & 57.94 \\
Diag. Fisher & 93.128 & 0.692 & 0.870 & 0.38 & 1.46 & 0.00 & 0.00 \\
\bottomrule
\end{tabular}
\caption{\textbf{Supplementary Zero Count comparison at $B=1{,}530$, including FT-800.} Table~1A uses Joint Frequency at $B=1{,}020$ for the same $n=256$. Here all eleven configurations use 1,024 sampled trials per model. All values are percentages. Target mass is $256P_i(i)$. Retain TV compares the conditional retain distribution with its uniform reference. Retrain TV compares the full model distribution with its paired retraining reference. Entries average three training seeds, except Max.\ FPR, the maximum over all target--control pairs and seeds. Emp.\ cov.\ applies the gates to observed rates. Supp.\ cov.\ applies them to simultaneous confidence bounds with tail allocation $0.05/[2(n^2+4n)]$.}
\label{tab:app-quality}
\end{table*}

Table~\ref{tab:app-quality} uses Zero Count with $B=1{,}530$ outputs per
trial and includes FT-800 alongside FT-80. Table~1A uses Joint Frequency
with $B=1{,}020$ at the same task size ($n=256$). Here Retraining meets the
simultaneous $90\%/5\%$ criterion for all 256 targets. We evaluate each
method on these targets with the same Zero Count rule. The coverage columns
use sampled outputs. The quality columns use the full model distributions.

For FT, increasing the edit budget from 80 to 800 updates reduces mean
normalized target mass from $58.120\%$ to $0.052\%$. Under this
supplementary protocol, both empirical and confidence-supported request
coverage rise from zero to $100\%$ for these two FT configurations.

The quality columns help interpret coverage alongside target removal and
retained-distribution behavior.
Original has zero empirical coverage and retains $100\%$ of normalized
target mass, even though its mean TV distance to Retraining is only
$0.395\%$. NegGrad+, NPO+R, and NPO+cKL reach $97.79$--$99.61\%$
empirical coverage, while their conditional retain TV is $2.11$--$2.50\%$,
compared with $0.027\%$ for Retraining. Thus a high coverage value alone
does not establish that the edited distribution preserves the retained
records' relative probabilities.

\end{document}